%% file: acl_latex.tex
\documentclass[11pt]{article}

\usepackage[final]{acl}

\usepackage{times}
\usepackage{latexsym}

\usepackage[T1]{fontenc}

\usepackage[utf8]{inputenc}

\usepackage{microtype}

\usepackage{inconsolata}

\usepackage{graphicx}

\usepackage{amsmath}
\usepackage{amssymb}
\usepackage{amsfonts}
\usepackage{booktabs}
\usepackage{multicol}
\usepackage{multirow}
\usepackage{bm}

\newcommand{\pldiff}[1]{\!\colorbox[cmyk]{0.1,0,0.1,0}{#1}\!}

\definecolor{green_mention}{HTML}{38761D}
\definecolor{orange_mention}{HTML}{f29637}
\definecolor{gray_mention}{HTML}{666666}

\usepackage{CJKutf8}
\usepackage{subcaption}
\usepackage{fontawesome}
\usepackage{tcolorbox}

\title{Recognizing Dialogue Ambiguity via Candidate Set Visual Grounding}

\author{
 \textbf{Shun Inadumi \textsuperscript{1}}, 
 \textbf{Lee Sangmyeong \textsuperscript{2,3}}, 
 \textbf{Koichiro Yoshino \textsuperscript{1,2,3}}
\\
 \textsuperscript{1} The University of Osaka, \\
 \textsuperscript{2} Guardian Robot Project, RIKEN,
 \textsuperscript{3} Nara Institute of Science and Technology
\\
 \texttt{inadumi@sys.es.osaka-u.ac.jp} \\
    \makebox[1.5em][c]{\faGithub}\url{https://github.com/OU-sigma-dis/candvg}\\
    \makebox[1.5em][c]{%
        \raisebox{-0.75ex}{\includegraphics[height=1.2em]{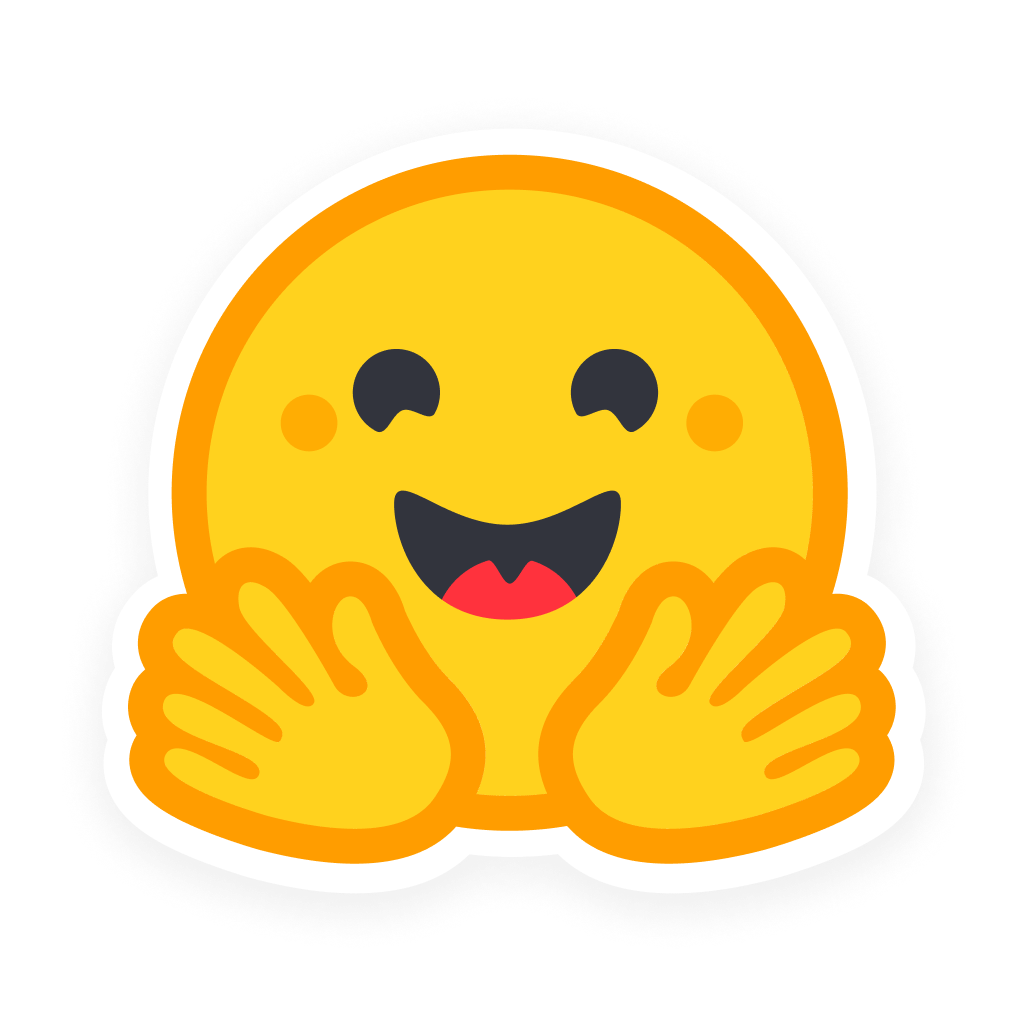}}%
    }\url{https://huggingface.co/datasets/SInadumi/CandVG}
}

\begin{document}
\maketitle
\begin{abstract}
Interpreting an ambiguous utterance in real-world interactions often requires context beyond a single utterance.
When the utterance refers to objects in a visual scene, it is necessary to maintain and update the set of plausible referents as context accumulates, without assuming that the available information identifies a unique object.
However, prior work on interactive visual grounding, which localizes objects referred to by linguistic expressions, has focused on identifying a unique object once sufficient dialogue context becomes available.
To address this gap, we propose Candidate Set Visual Grounding (CSVG), a task for estimating the set of objects consistent with the given dialogue context.
We further introduce CandVG, a benchmark that evaluates how the candidate set changes across dialogue turns.
Our evaluation of vision-language models shows that stronger visual grounding capability does not necessarily yield better CSVG performance.
In particular, models have difficulty predicting the complete candidate set, owing to challenges in candidate generation and premature commitment to a small set of candidates in early turns.
Nevertheless, we find that this limitation can be partially mitigated in the evaluated large-scale models using instructions designed for CSVG.

\end{abstract}

\input{contents/1_introduction}
\input{contents/2_related_work}
\input{contents/3_problem_definition}
\input{contents/4_benchmark}
\input{contents/5_experiments}

\input{contents/6_analysis}
\input{contents/7_conclusion}

\section*{Limitations}

\paragraph*{Benchmark.}
CandVG is constructed from the existing dataset, and its image domains, dialogue styles, and ambiguity patterns therefore reflect those of the source data.
In most dialogues, candidate sets gradually shrink as additional context becomes available, often converging to a single object.
Although a small number of dialogues involve sudden shifts in the intended object, resulting in candidate addition, such transition patterns are underrepresented in our benchmark.
Future work should extend CandVG with more diverse dialogue structures and candidate set transitions.

\paragraph*{Setting.}
We consider static images and represent candidate objects using BBoxes.
This setting does not capture changes in the environment, viewpoint-dependent references, or non-verbal cues that may arise during real-world interaction.
A single BBox may not precisely represent certain referents, such as object parts or regions.
Extending our work to dynamic, embodied, and potentially real-world environments is therefore an important direction.

\paragraph*{Model.}
Our evaluation covers a diverse set of recent VLMs, but the open-weight models are limited to at most 38B parameters because of available computational resources, while the proprietary evaluation excludes several major model families, including Gemini~\cite{Gemini_2025_genimi-robotics}.
The observed scaling trends may therefore not generalize to larger open-weight models or future model families.
Broader evaluation is needed to assess the generality of our findings across model architectures, sizes, and training objectives.

\paragraph*{Limited Improvement Strategies.}
Our investigation of performance improvement is limited to task-aware instructions applied at inference time.
We have not systematically evaluated more sophisticated prompting or decoding strategies, such as self-reflection~\cite{Noah-etal_2023_NeurIPS}.

Although task-aware instructions improve recall for some large-scale VLMs, they do not yield consistent gains for compact VLMs.
Such compact models are preferable for deployment in interactive systems because of their lower computational and latency requirements.
Future work should therefore explore training and inference methods specifically designed to improve candidate set estimation in compact models and facilitate their deployment in real-world interactions.

\section*{Ethical Consideration}
CandVG builds on an existing CVG dataset~\cite{Zhang-etal_2024_CVPRW} and is designed for research purposes.
The underlying images come from the Open Images Dataset~\cite{Kuznetsova-etal_2020_OpenImages} and are provided under Creative Commons Attribution licenses; Open Images annotations are released under CC BY 4.0.
We will release the newly collected CandVG annotations under CC BY 4.0.
We note that some images contain people and that CandVG may inherit representational biases present in the source datasets.
The annotators were proficient in English and able to understand the dialogues used for candidate set annotation.
Annotation contributors were compensated at JPY 2,000 per hour.

\section*{Acknowledgments}
This work was supported by RIKEN Junior Research Associate Program and JST, PRESTO Grant Number JPMJPR24TC, Japan.
A part of this research was supported by FRONTia, a Japanese national program led by the Ministry of Economy, Trade and Industry of Japan (METI) and the New Energy and Industrial Technology Development Organization (NEDO), aimed at developing a domestic multimodal foundation model for AI robots and physical AI.
\bibliography{custom}

\appendix
\input{contents/appendix}

\end{document}

%% file: contents/1_introduction.tex
\begin{figure}[t]
    \centering
    \includegraphics[keepaspectratio, width=0.48\textwidth]{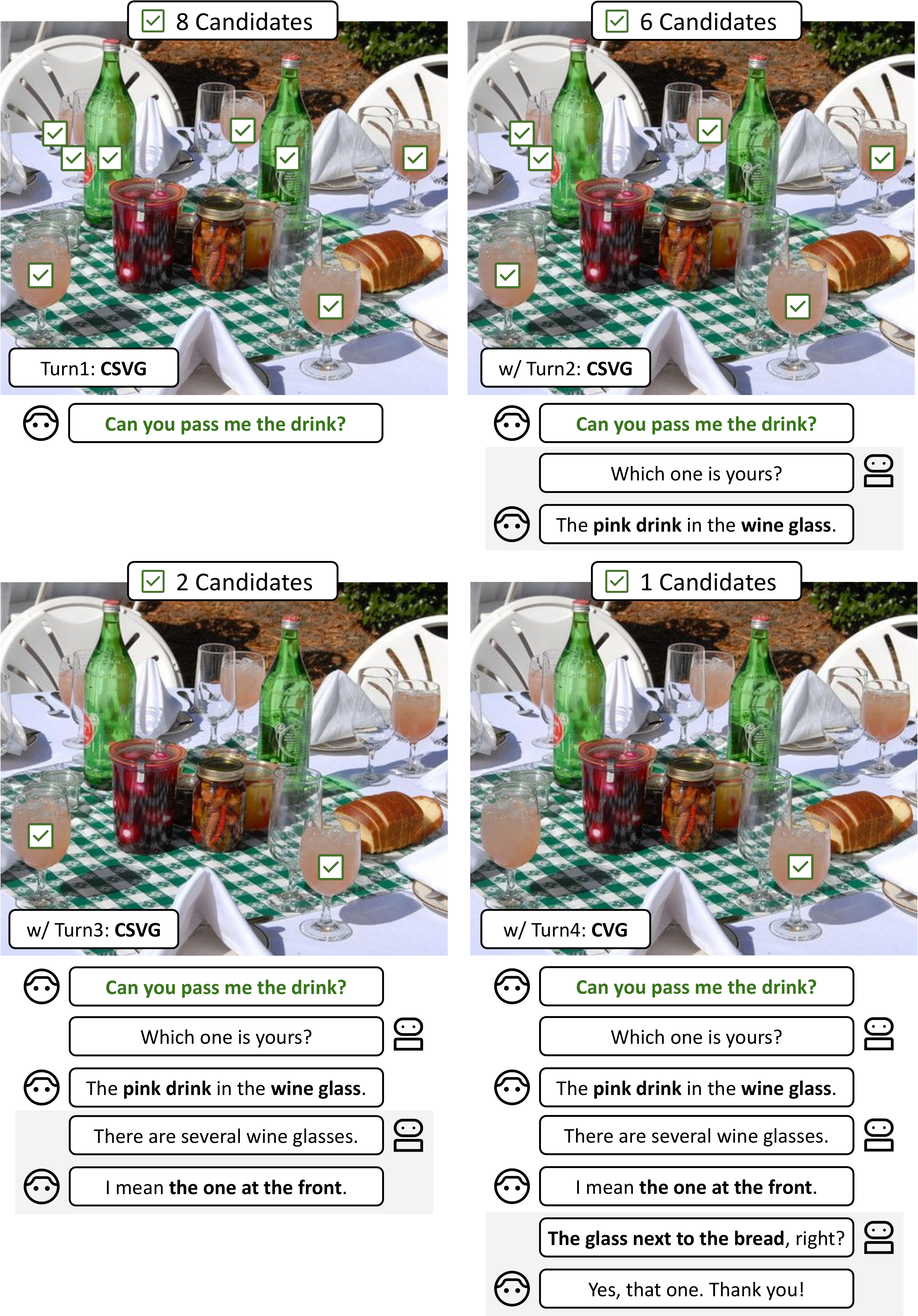}
    \caption{Concept of CandVG: Checkboxes indicate objects that remain plausible referents of the \textcolor[HTML]{37761D}{initial ambiguous utterance} under the given dialogue context. Each accumulated turn yields a separate evaluation sample.}
    \label{fig:1_candvg}
\end{figure}

\section{Introduction}
In real-world interactions, interpreting an utterance often requires context that is not explicitly stated in the utterance itself~\cite{Clark_1996_using-language, Piantadosi-etal_2012}.
For example, in Figure~\ref{fig:1_candvg}, the initial utterance ``\emph{Can you pass me the drink?}'' cannot be uniquely interpreted from the utterance and visual scene alone, requiring additional context to identify the intended object.
Appropriately handling such ambiguity depending on context is essential for developing intelligent systems that can collaborate with users in the real world~\cite{Tellex-etal_2011_AAAI, Lemaignan-etal_2012_grounding-interaction}.

When an utterance refers to objects in a visual scene, this problem is commonly formulated as visual grounding, the task of localizing the referred objects~\cite{Kazemzadeh-etal_2014_EMNLP}.
In user--system interactions, prior work has used dialogue context to resolve ambiguity about which object a user's utterance refers to~\cite{Vries-etal_2017_CVPR, Alamri-etal_2019_CVPR}.
We refer to this formulation in visual grounding as Context-Aware Visual Grounding (CVG).
CVG assumes sufficient context to identify the targeted object; however, real-world interactions also include cases where the target cannot be narrowed down based solely on the given context.

These cases require systems to avoid prematurely committing to a single object and instead maintain hypotheses about the set of possible candidates, providing a basis for downstream actions such as asking clarifying questions~\cite{Kruijff-etal_2006_HRI, Shridhar-and-Hsu_2018_RSS}.
Recent work has introduced problem formulations that allow systems to identify multiple candidates for ambiguous linguistic expressions without assuming a single interpretation~\cite{Chen-etal_2025_ICCV, Zhu-etal_2026_ICML}.
However, it remains underexplored how to evaluate the candidate set itself, particularly \emph{whether systems can track how it changes as additional context becomes available}.

To address this, we propose Candidate Set Visual Grounding (CSVG), a task for recognizing unresolved ambiguity by identifying all candidates that remain plausible under the given context.
We further introduce CandVG, a benchmark that evaluates how the candidate set changes as dialogue context accumulates.
As illustrated in Figure~\ref{fig:1_candvg}, CandVG captures how the candidate set narrows across dialogue turns, where the turn yielding a unique candidate corresponds to the CVG task.
It consists of 1,249 dialogues and 4,833 evaluation samples.
Each accumulated context is treated as a separate sample for identifying the set of candidates referred to by the initial ambiguous utterance.

We evaluate 18 vision-language models (VLMs) across six families, covering both open-weight and proprietary VLMs.
They provide a strong baseline for assessing whether current systems can recognize unresolved ambiguity in dialogue, given their potential to use the context for visual grounding~\cite{You-etal_2024_Ferret, Rasheed-etal_2024_CVPR_GLaMM}.
We evaluate the models on CandVG and standard visual grounding benchmarks~\cite{Mao-etal_2016_CVPR, Bryan-etal_2017_IJCV_Flickr30k, He-etal_2023_gRefCOCO} to examine how standard visual grounding ability relates to CVG and CSVG performance.

Our experiments show that larger VLMs generally achieve higher CVG performance, but increased model scale does not necessarily lead to better CSVG performance ({\S}~\ref{sec:5_main_results_candvg}).
In particular, models tend to predict only part of the candidate set, while missing many others in CSVG ({\S}~\ref{sec:5_main_results_candvg} and {\S}~\ref{sec:5_comparison-vg}).
This limitation is attributed to two tendencies in the model.
First, models have difficulty with candidate generation, with performance improving when ground-truth candidates are provided ({\S}~\ref{sec:5_effect-gt-cand}).
Second, some models exhibit premature commitment by narrowing the candidate set aggressively in early turns while the utterance remains ambiguous ({\S}~\ref{sec:6_tracking_candidate_set}).
Building on these observations, we find that prompts designed for CSVG partially mitigate this limitation, although the effects remain limited and model-dependent ({\S}~\ref{sec:6_task-aware-prompting}).
These findings provide insights into how VLMs can recognize all plausible referents of an ambiguous utterance by tracking changes in dialogue context.


%% file: contents/2_related_work.tex
\section{Related Work}
Visual grounding is the task of localizing objects in an image referred to by a linguistic expression~\cite{Kazemzadeh-etal_2014_EMNLP, Xiao-etal_2025_TPAMI}.\footnote{Depending on the type of linguistic expression, this task is also referred to as referring expression comprehension~\cite{Mao-etal_2016_CVPR} or phrase grounding~\cite{Bryan-etal_2017_IJCV_Flickr30k}.}
This task supports a range of applications involving language understanding in visual scenes, including visual question answering~\cite{Antol-etal_2015_ICCV_VQA, Zhu-etal_2016_CVPR}, vision--language navigation~\cite{Anderson-etal_2018_CVPR_VLN, Liu-etal_2025_WACV}, and user--system interactions~\cite{Shridhar-and-Hsu_2018_RSS}.
From this perspective, we study utterance ambiguity in dialogue as a visual grounding problem, focusing on which objects an utterance may refer to.

In user--system interactions, interpreting an ambiguous utterance often requires context beyond the utterance itself.
Prior work on interactive visual grounding has therefore used dialogue context to resolve ambiguity about which object the utterance refers to~\cite{Vries-etal_2017_CVPR, Kottur-etal_2022_EMNLP_SIMMC, Ueda-etal_2024_LREC-COLING, Zhang-etal_2024_CVPRW}.
We also focus on dialogue context and call this task Context-Aware Visual Grounding (CVG).

Typical visual grounding approaches can be broadly categorized into object detector-based~\cite{Kamath-etal_2021_ICCV_MDETR, Li-etal_2022_CVPR_GLIP, Liu-etal_2024_ECCV_GDino} and VLM-based~\cite{Chen-etal_2023_miniGPTv2, Peng-etal_2024_ICLR_Cosmos2, Hao-etal_2024_ECCV_LLaVAGrounding, Chen-etal_2024_CVPR_InternVL, Bai-etal_2025_Qwen3VL}.
Object detector-based approaches can offer fast inference~\cite{Fu-etal_2026_CVPR_WeDetect}, whereas VLM-based approaches can flexibly interpret diverse linguistic expressions~\cite{Lai-etal_2024_CVPR_LISA} and incorporate additional textual context~\cite{Chen-etal_2023_Shikra, You-etal_2024_Ferret, Rasheed-etal_2024_CVPR_GLaMM}.
Accordingly, we focus on VLM-based approaches, as they are well suited to visual grounding tasks that require context, such as CVG.

Although CVG assumes that context is sufficient to identify a single object, multiple candidates may remain plausible under the available context.
In practical interactive systems, interpreting ambiguous user utterances typically involves first identifying candidates and then asking clarification questions~\cite{Deits-etal_2013_JHRI, Shridhar-and-Hsu_2018_RSS, Hatori-etal_2018_ICRA, Dogan-etal_2022_HRI, Mo-etal_2023_ICRA}.
Recent work has begun to ground the interpretation of linguistic expressions in visual scenes, studying answerability in 3D question answering~\cite{Azuma-etal_2024_IROS} and unresolved ambiguity in both visual question answering~\cite{Chen-etal_2025_ICCV} and 3D visual grounding~\cite{Zhu-etal_2026_ICML}.
Building on these lines of work, we propose Candidate Set Visual Grounding (CSVG) to evaluate whether systems can identify the set of candidates that remain plausible given the available context, and introduce CandVG as a benchmark that covers both CVG and CSVG.\footnote{Unlike existing CVG datasets, which assume contexts sufficient to identify a single referent, CandVG also includes contexts in which multiple referents remain plausible.}


%% file: contents/3_problem_definition.tex
\section{Problem Formulation}
\label{sec:3_problem_definition}

\subsection{Task Setting}
Given an RGB image $\mathbf{I} \in \mathbb{R}^{H \times W \times 3}$ and a dialogue context $\mathbf{T}_i = (t_1, \ldots, t_i)$ available through the $i$th dialogue turn, our goal is to identify the objects that remain plausible referents of the utterance $t_1$ under $\mathbf{T}_i$. 
Here, $H$ and $W$ denote the image height and width, respectively.\footnote{In CandVG, $t_1$ corresponds to the initial ambiguous utterance, while each subsequent turn consists of a system question and a user response.}
The output is a candidate set of bounding boxes (BBoxes) $\mathcal{B}$ corresponding to the identified objects.

We define two tasks according to the size of $\mathcal{B}$:
\begin{description}
    \item[CVG: Context-Aware Visual Grounding.] \leavevmode\\
    The context $\mathbf{T}_i$ is sufficient to identify a single object, and the goal is to predict its BBox.
    \item[CSVG: Candidate Set Visual Grounding.] \leavevmode\\
    Multiple objects remain plausible as referents of $t_1$ under the context $\mathbf{T}_i$, and the goal is to predict their BBoxes.
\end{description}

\subsection{Evaluation Metrics}
\label{sec:eval_metrics}
We evaluate CVG using Acc@0.5 (Acc), a standard metric for visual grounding~\cite{Xiao-etal_2025_TPAMI}.
A prediction is counted as correct if its Intersection over Union (IoU) with the ground-truth BBox is at least 0.5.
When multiple BBoxes are returned, we consider only the first as the prediction.

CSVG evaluates the correspondence between the predicted and ground-truth sets of BBoxes.
For each sample, we first compute the pairwise IoUs between predicted and ground-truth BBoxes and then perform one-to-one matching using the Hungarian algorithm to maximize the number of matched pairs with IoU at least 0.5.
Precision (P), recall (R), and F1 score are computed from the resulting matches as follows:
\begin{align*}
\mathrm{P} = \frac{\mathrm{TP}}{\mathrm{TP} + \mathrm{FP}}, 
\mathrm{R} = \frac{\mathrm{TP}}{\mathrm{TP} + \mathrm{FN}},
\mathrm{F1} = \frac{2 \cdot \mathrm{P} \cdot \mathrm{R}}
{\mathrm{P} + \mathrm{R}},
\end{align*}
where TP, FP, and FN denote matched pairs, unmatched predicted BBoxes, and unmatched ground-truth BBoxes, respectively.
We compute these metrics for each sample and report their averages over all samples.

%% file: contents/4_benchmark.tex
\section{CandVG}
\label{sec:4_candvg_benchmark}
We constructed CandVG, a benchmark covering both CVG and CSVG under a dialogue context.
CandVG is built upon an existing CVG dataset containing natural human--human dialogues and BBox annotations~\cite{Zhang-etal_2024_CVPRW}.
For each context $\mathbf{T}_i$, annotators identified the plausible candidates using point-based annotations, which were then mapped to the corresponding BBoxes to construct a single ground-truth candidate set containing all plausible referents.
This process yielded one evaluation sample per context, resulting in 1,249 dialogues and 4,833 evaluation samples.

\subsection{Data Collection}
\label{sec:4_candvg_data_collection}
Using dialogue and BBox annotations from an existing dataset, we annotate the candidate set for each dialogue context by placing points on all plausible objects.
The annotation process consists of two stages, both performed by three annotators assigned equal portions of the dialogues:
\begin{description}
    \item[Data Selection.]
    We automatically sampled 1,620 dialogues to balance the distribution of candidate set sizes.
    We then manually screened these dialogues to ensure benchmark quality, excluding 303 dialogues with annotation errors or no ambiguity in the first utterance.
    This process retained 1,317 dialogues.

    \item[Candidate Set Collection.]
    For each selected dialogue, we annotated the candidate set for each dialogue context.
    Annotators first placed one point on each plausible object and then updated the candidate set at subsequent turns by adding or removing points.\footnote{Points annotated for the first utterance were carried over to subsequent turns because later contexts typically refine the candidate set. Annotators could nevertheless add new points when the updated context made additional objects plausible.}
\end{description}
After annotation, we mapped each point to its corresponding BBox provided by the existing CVG dataset~\cite{Zhang-etal_2024_CVPRW} and retained only dialogues with a complete one-to-one correspondence between points and BBoxes.
The resulting CandVG benchmark comprises 1,249 dialogues.
Appendix~\ref{sec:appendix_data_collection_details} describes the data collection criteria and procedures in detail.

\begin{figure}[t]
    \centering
    \begin{subfigure}{\linewidth}
        \includegraphics[keepaspectratio, width=\linewidth]{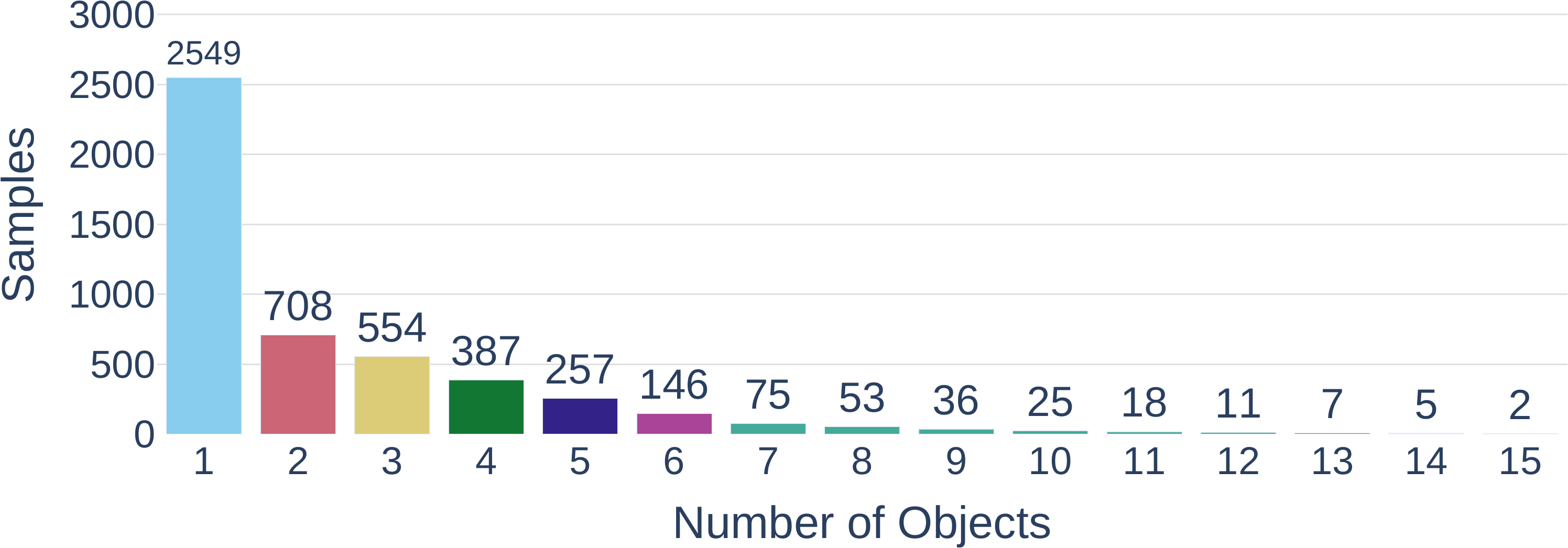}
        \subcaption{Candidate set size across all samples}
        \label{fig:4_object_dist}
    \end{subfigure}
    \vspace{0.5em}
    \begin{subfigure}{\linewidth}
        \centering
        \includegraphics[keepaspectratio, width=\linewidth]{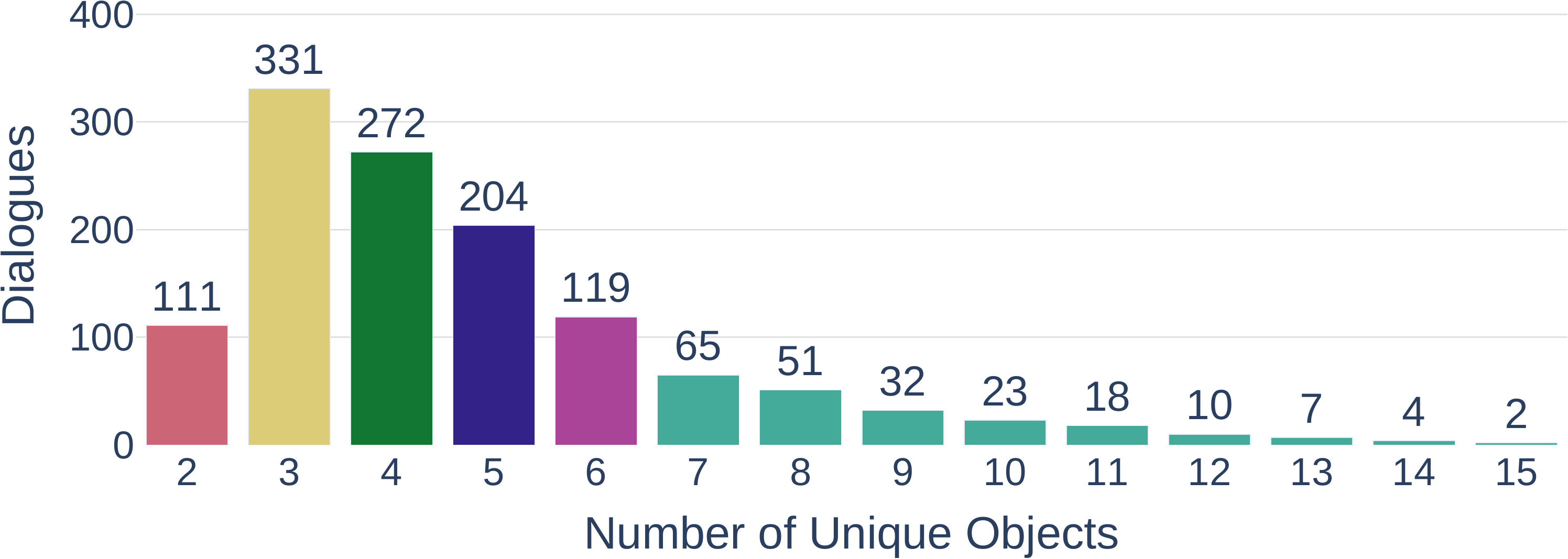}
        \subcaption{Unique candidates across each dialogue}
        \label{fig:4_uniq_object_dist_per_dialogue}
    \end{subfigure}
    \caption{Distributions of candidate set size per sample and number of unique candidates across each dialogue}
    \label{fig:4_distribution}
\end{figure}

\subsection{Statistics and Analysis}
\paragraph*{Object Distribution.}
Figure~\ref{fig:4_distribution} shows the distribution of candidate set sizes across all evaluation samples and the number of unique candidates across each dialogue.
In particular, Figure~\ref{fig:4_distribution} (b) reflects the number of potential distractor objects in CVG and CSVG.
CVG comprises 2,549 samples with a single object under the available context, whereas the remaining 2,284 samples with multiple objects constitute CSVG.
Additional statistics and analysis of CandVG are provided in Appendix~\ref{sec:appendix_additional_stats_analysis}.

\paragraph*{Data Verification.}
To verify annotation quality, we measure inter-annotator agreement using the CVG and CSVG metrics, as identifying candidate sets through point annotations involves human judgment.
We sampled 30 dialogues from each of six bins (2, 3, 4, 5, 6, and 7+) to cover different numbers of unique candidates, as shown in Figure~\ref{fig:4_distribution} (b).
Two annotator pairs each independently annotated a separate set of 180 dialogues, totaling 360 dialogues.
For each pair, we report the CVG and CSVG evaluation metrics separately.
Table~\ref{tab:iaa} shows high agreement for both annotator pairs across the CVG and CSVG metrics, supporting the reliability of the CandVG annotations.

\begin{table}[t]
    \centering
    \scalebox{0.86}{
    \begin{tabular}{l|c|ccc}
        \toprule
        \textbf{Pair} & \textbf{Acc} & \textbf{P} & \textbf{R} & \textbf{F1} \\
        \midrule
        Pair1 & 0.962 & 0.959 & 0.944 & 0.943 \\
        Pair2 & 0.971 & 0.966 & 0.924 & 0.935 \\
        \bottomrule
    \end{tabular}
    }
    \caption{Annotation agreement for two annotator pairs}
    \label{tab:iaa}
\end{table}

%% file: contents/5_experiments.tex
\section{Experiments}
\label{sec:5_candvg_experiments}

\subsection{Settings}
\paragraph*{Compared Models.}
We evaluated open-weight and proprietary generalist VLMs that can directly generate BBoxes in a zero-shot setting.
This setting allowed us to assess whether models can use dialogue context to identify the objects referred to by the first ambiguous utterance without task-specific fine-tuning.
Specifically, we evaluated the open-weight InternVL3.5~\cite{Wang-etal_2025_InternVL35} (Intern3.5), Qwen3 VL~\cite{Bai-etal_2025_Qwen3VL}, and Qwen3.5~\cite{QwenTeam_2026_Qwen35} families, together with the proprietary GPT-5.6~\cite{OpenAI_2026_GPT-5.6}.
To examine the effect of model scale, we included variants of the open-weight models ranging from 2B to 38B parameters.

In addition to generalist VLMs, we evaluated embodied reasoning VLMs based on Qwen3 VL and post-trained for spatiotemporal and physical reasoning in embodied scenarios.
For this comparison, we included the RynnBrain~\cite{Dang-etal_2026_RynnBrain} and CosmosReason2~\cite{NVIDIA_2026_CosmosReason2} families to examine whether post-training for embodied reasoning improves performance on CVG and CSVG.

\paragraph*{Evaluation Benchmarks.}
In addition to CandVG, we evaluated models on standard visual grounding benchmarks to separate general visual grounding ability from the ability to identify candidate sets using context.
In the CVG setting, we used RefCOCOg~\cite{Mao-etal_2016_CVPR} and Flickr30k Entities (F30k-Ent)~\cite{Bryan-etal_2017_IJCV_Flickr30k}, both of which evaluate the localization of a single object.
For each benchmark, we used the same number of evaluation samples as in the CandVG CVG evaluation set.\footnote{Flickr30k Entities is a phrase grounding benchmark. We use only samples in which a phrase refers to a single object.}
The CSVG setting uses gRefCOCO~\cite{He-etal_2023_gRefCOCO}, which evaluates the localization of multiple objects referred to by a referring expression.\footnote{RefCOCOg and gRefCOCO are distinct benchmarks; the former is used for single-object visual grounding, while the latter is used for multi-object visual grounding.}
To make the evaluation set comparable to CandVG, we matched both the total number of samples and the distribution of candidate set sizes across bins 2, 3, 4, 5, 6, and 7+.

\paragraph*{Candidate Set Generation and Selection.}
We considered the CSVG task in CandVG under two conditions to assess how the availability of ground-truth candidates affects the F1 score.
In the \emph{without ground-truth candidates (w/o GT)} condition, models predicted the candidate set directly from the image and context, whereas in the \emph{with ground-truth candidates (w/ GT)} condition, the ground-truth candidate set was additionally provided as input.
This setting enabled a direct evaluation of candidate set estimation under the given context while providing insight into the respective roles of candidate generation and candidate selection.

\paragraph*{Prompt Design.}
To ensure a fair comparison across benchmarks, we used the same instruction for all models.
We asked models to output BBoxes and limit the number of outputs to 15 to match the maximum candidate set size in CandVG:
\begin{tcolorbox}[colframe=gray,fontupper=\scriptsize,title=\small Base Prompt]
\small
Provide the bounding box coordinates of the text described between <ref> and </ref>.

Return as box(es) as necessary, but include all plausible referents, up to a maximum of 15.
\end{tcolorbox}
We kept the base prompt and adapted only the BBox output format for each model, since supported formats differ across models.\footnote{We selected the model-specific output format based on performance on the
Flickr30k Entities validation split.}
We provide further implementation details in Appendix~\ref{sec:appendix_impl_details}.

\begin{table*}[t]
    \centering
    \begin{minipage}[t]{0.70\textwidth}
        \centering
        \vspace{0pt}
        \scalebox{0.825}{
        \begin{tabular}{l|ccc|cc}
            \toprule
            \multirow{2}{*}{\textbf{Models}}
            & \multicolumn{3}{c|}{\textbf{CVG Setting (Acc)}}
            & \multicolumn{2}{c}{\textbf{CSVG Setting (F1)}} \\
            & \makebox[1.5cm][c]{\textbf{CandVG}} & \makebox[1.5cm][c]{\textbf{RefCOCOg}} & \makebox[1.5cm][c]{\textbf{F30k-Ent}} & \makebox[1.5cm][c]{\textbf{CandVG}} & \makebox[1.5cm][c]{\textbf{gRefCOCO}} \\
            \midrule

            \addlinespace[2pt]
            \multicolumn{6}{c}{\textbf{Generalist VLMs}} \\
            \addlinespace[2pt]

            Intern3.5 2B & 0.521 & 0.859 & 0.666 & 0.329 & 0.419 \\
            Intern3.5 4B & 0.568 & 0.859 & 0.479 & 0.357 & \pldiff{0.674} \\
            Intern3.5 8B & 0.580 & 0.877 & \pldiff{0.757} & 0.331 & 0.496 \\
            Intern3.5 38B & \pldiff{0.746} & \pldiff{0.898} & 0.669 & \pldiff{0.382} & 0.646 \\

            \addlinespace[3pt]

            Qwen3 VL 2B & 0.480 & 0.852 & 0.519 & 0.271 & 0.391 \\
            Qwen3 VL 4B & 0.622 & 0.888 & 0.614 & \pldiff{0.542} & 0.560 \\
            Qwen3 VL 8B & 0.655 & 0.884 & 0.775 & 0.454 & 0.576 \\
            Qwen3 VL 32B & \pldiff{0.762} & \pldiff{0.910} & \pldiff{0.835} & \pldiff{0.543} & \pldiff{0.589} \\

            \addlinespace[3pt]

            Qwen3.5 2B & 0.606 & 0.856 & 0.504 & 0.410 & 0.507 \\
            Qwen3.5 4B & 0.735 & 0.888 & 0.670 & \pldiff{0.494} & \pldiff{0.672} \\
            Qwen3.5 9B & 0.790 & 0.899 & 0.794 & 0.427 & 0.510 \\
            Qwen3.5 27B & \pldiff{\textbf{0.819}} & \pldiff{\textbf{0.917}} & \pldiff{\textbf{0.854}} & 0.441 & 0.497 \\

            \addlinespace[3pt]

            GPT-5.6 Luna & 0.743 & 0.844 & 0.807 & \pldiff{\textbf{0.649}} & 0.721 \\
            GPT-5.6 Terra & \pldiff{0.780} & \pldiff{0.863} & \pldiff{0.814} & \pldiff{\textbf{0.649}} & \pldiff{\textbf{0.763}} \\

            \midrule

            \addlinespace[2pt]
            \multicolumn{6}{c}{\textbf{Embodied Reasoning VLMs} based on Qwen3 VL} \\
            \addlinespace[2pt]

            RynnBrain 2B & 0.599 & 0.837 & 0.520 & 0.342 & \pldiff{0.277} \\
            RynnBrain 8B & \pldiff{0.691} & \pldiff{0.865} & \pldiff{0.705} & \pldiff{0.356} & 0.256 \\
            CosmosReason2 2B & \pldiff{0.537} & 0.836 & 0.548 & 0.326 & 0.494 \\
            CosmosReason2 8B & 0.519 & \pldiff{0.868} & \pldiff{0.741} & \pldiff{0.478} & \pldiff{0.497} \\

            \bottomrule
        \end{tabular}
        }
        \captionof{table}{
            Overall results on CandVG and standard visual grounding
            benchmarks: \pldiff{Colored} and \textbf{bold} values denote the best results
            \pldiff{within each model family} and \textbf{across all models}, respectively.
        }
        \label{tab:4_overall_results}
    \end{minipage}
    \hfill
    \begin{minipage}[t]{0.288\textwidth}
        \centering
        \vspace{0pt}
        \begin{minipage}[t]{\linewidth}
            \centering
            \includegraphics[width=\linewidth]{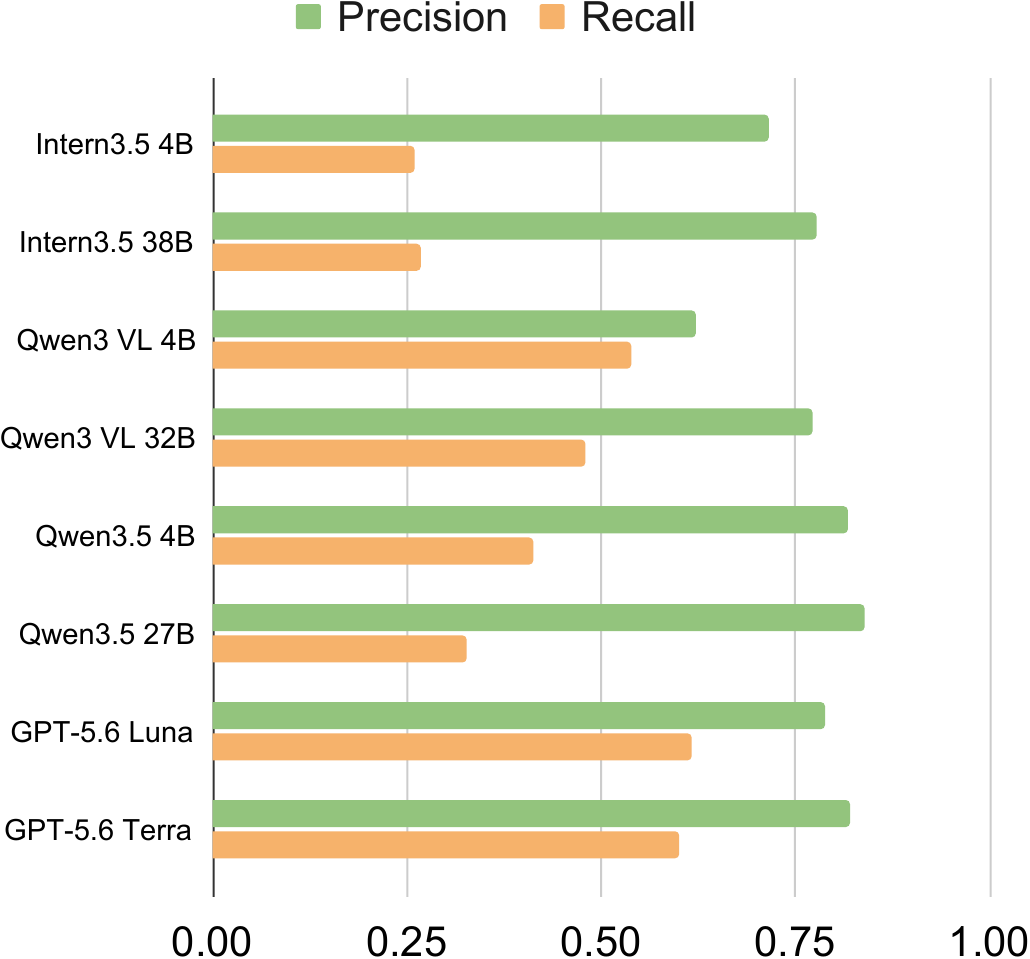}
            \subcaption{CandVG}
            \label{fig:csvg_prec-rec}
        \end{minipage}
        
        \vspace{3.1mm}
        
        \begin{minipage}[t]{\linewidth}
            \centering
            \includegraphics[width=\linewidth]{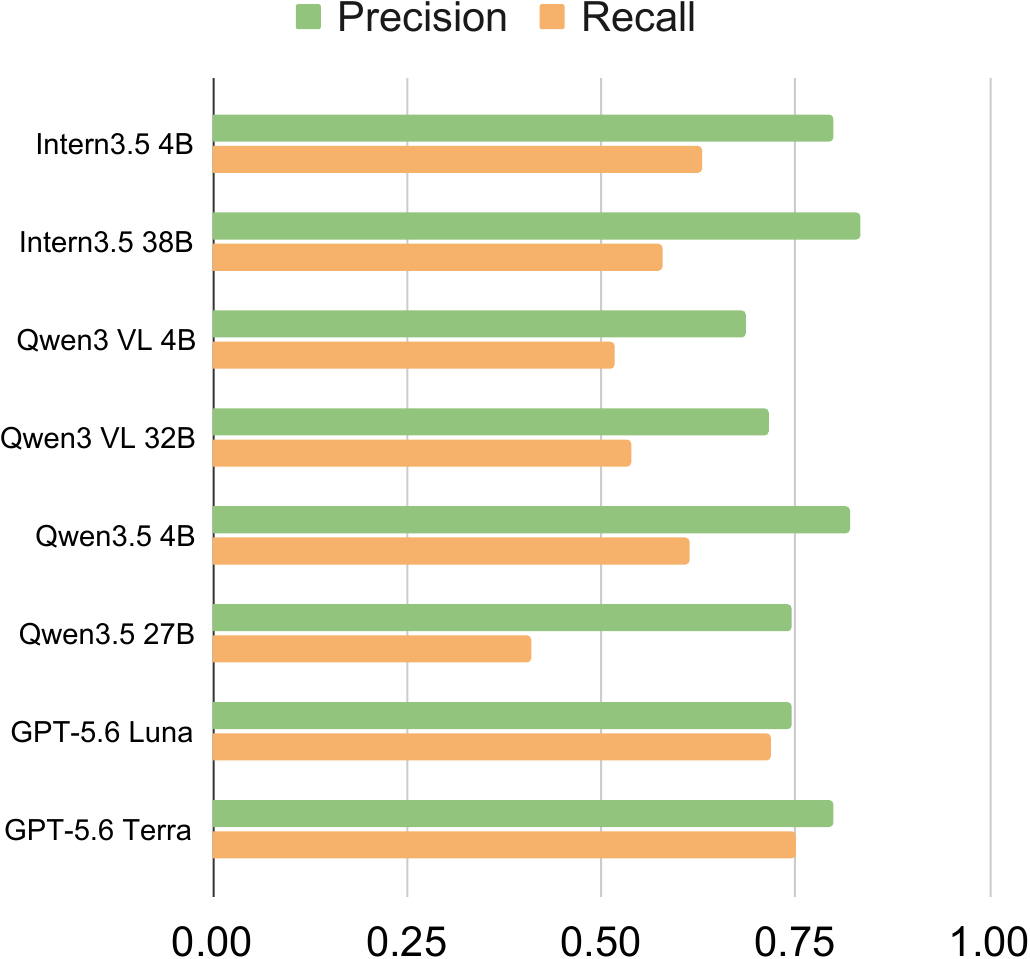}
            \subcaption{gRefCOCO}
            \label{fig:grefcoco_prec-rec}
        \end{minipage}
        \captionof{figure}{Precision and recall in CSVG setting}
        \label{fig:csvg_precision_recall}
    \end{minipage}
\end{table*}

\subsection{Main Results}
\label{sec:5_main_results_candvg}
Table~\ref{tab:4_overall_results} reports CVG accuracy and CSVG F1 score. 
Figure~\ref{fig:csvg_precision_recall} reports CSVG precision and recall for 4B and 27B--38B open-weight generalist VLMs and both GPT-5.6 variants.

\paragraph*{Scaling Trends in CVG and CSVG.}
Table~\ref{tab:4_overall_results} shows a scaling trend for the CVG task on CandVG, with accuracy increasing consistently with model scale within each open-weight generalist VLM family.
Among all evaluated models, Qwen3.5 27B achieves the highest accuracy.
These results indicate that \emph{larger generalist VLMs make better use of dialogue context in the CVG task}.

For the CSVG task on CandVG, Table~\ref{tab:4_overall_results} shows that larger open-weight generalist VLMs do not consistently achieve higher F1 scores.
Across all evaluated model families, GPT-5.6 achieves the highest F1 score, while the best-performing open-weight models come from the Qwen3 VL family.
In contrast to CVG, \emph{increasing model scale alone does not consistently improve candidate set estimation on the CSVG task}.

\paragraph*{Precision--Recall Gap in CSVG.}
Figure~\ref{fig:csvg_precision_recall} (a) shows a precision--recall gap for the CSVG task on CandVG, with recall consistently lower than precision.
Specifically, precision increases with model scale, whereas recall does not improve for larger models.
This pattern shows that models can identify precise subsets of the ground-truth candidate sets but \emph{fail to estimate all plausible candidates under the given context}.

\paragraph{Generalist vs. Embodied Reasoning VLMs.}
Table~\ref{tab:4_overall_results} shows that embodied reasoning post-training does not yield consistent gains across CandVG.
Both 2B embodied reasoning models outperform Qwen3 VL 2B on the CVG and CSVG tasks.
However, their 8B variants show opposite patterns relative to Qwen3 VL 8B: RynnBrain performs better on CVG but worse on CSVG, whereas CosmosReason2 shows the reverse.

\subsection{Comparison with Standard Visual Grounding Benchmarks}
\label{sec:5_comparison-vg}
Focusing on the CVG setting in Table~\ref{tab:4_overall_results}, \emph{models with stronger visual grounding performance achieve higher accuracy on CandVG}.
Among the evaluated models, Qwen3.5 27B obtains the highest score across all benchmarks in the CVG setting.
At the same time, all models score lower on CandVG than on RefCOCOg, and most also score lower than on F30k-Ent.
This lower accuracy may be attributed to the need to account for dialogue context when interpreting an ambiguous utterance.

Table~\ref{tab:4_overall_results} shows that VLMs achieve higher F1 scores on gRefCOCO than on CandVG in the CSVG setting.
Figure~\ref{fig:csvg_precision_recall} further shows that, although both benchmarks exhibit a precision--recall gap, \emph{the gap is consistently larger on CandVG}.
Among the open-weight generalist VLMs, Qwen3 VL 4B shows the smallest gap between the two benchmarks.
This comparison highlights the challenge of CandVG beyond standard multi-object grounding, as models need to infer which candidates remain plausible using the context.

\begin{figure}[t]
    \centering
    \includegraphics[width=\linewidth]{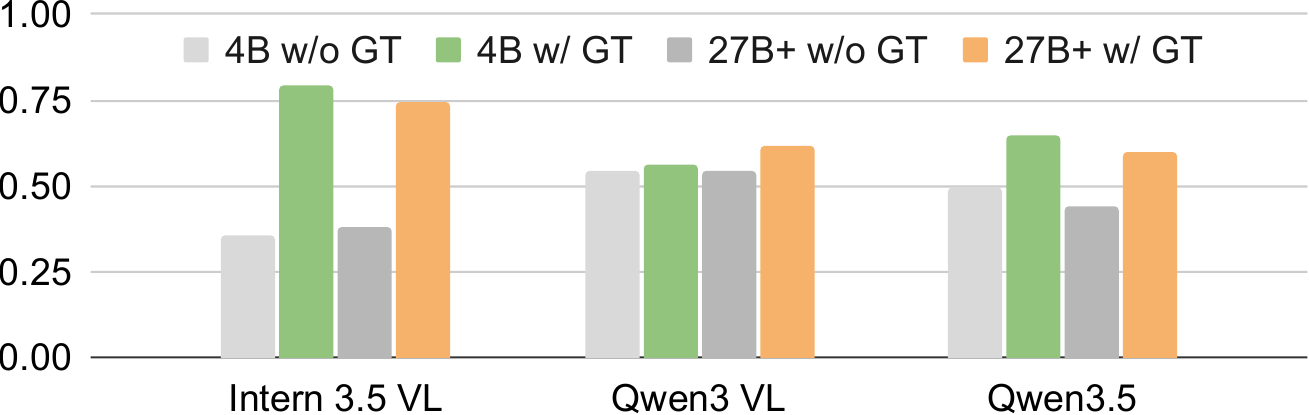}
    \caption{F1 scores on CSVG with and without ground-truth candidate sets, where 27B+ denotes the 27B--38B variants.}
    \label{fig:candvg_gt}
\end{figure}

\subsection{Effect of Ground-Truth Candidates}
\label{sec:5_effect-gt-cand}
To examine whether the low recall in the CSVG task primarily arises from candidate generation, we provide ground-truth candidate sets to 4B and 27B--38B open-weight generalist VLMs.
As shown in Figure~\ref{fig:candvg_gt}, all models achieve higher F1 scores in the w/ GT condition than in the w/o GT condition.
In particular, the gains are large for Intern3.5 variants.
At the same time, larger VLMs do not consistently achieve higher F1 scores even under the w/ GT condition.
These results indicate that \emph{candidate generation is one bottleneck in the CSVG task}, as providing ground-truth candidates substantially reduces this burden.

%% file: contents/6_analysis.tex
\begin{figure*}[t]
    \centering
    \begin{subfigure}[t]{0.32\linewidth}
        \centering
        \includegraphics[width=\linewidth]{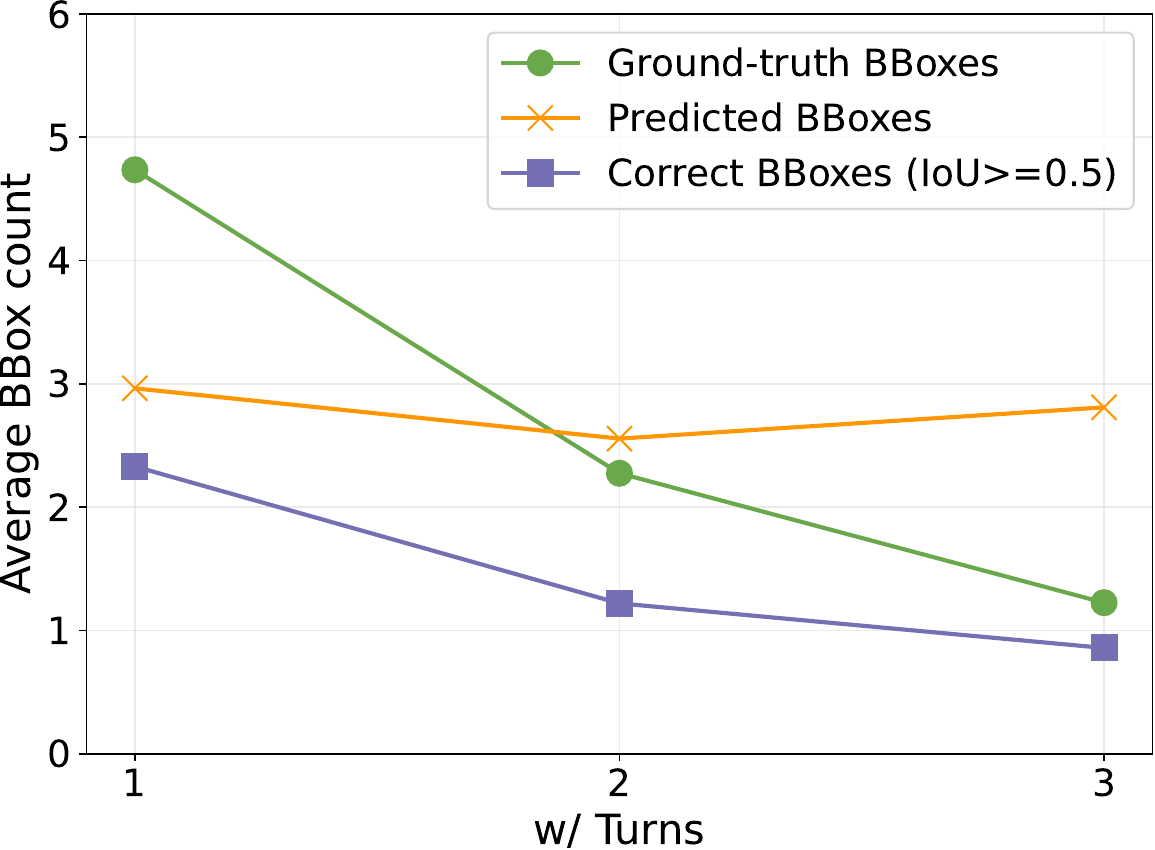}
        \subcaption{Qwen3 VL 4B}
        \label{fig:ct_qwen3-4B}
    \end{subfigure}
    \begin{subfigure}[t]{0.32\linewidth}
        \centering
        \includegraphics[width=\linewidth]{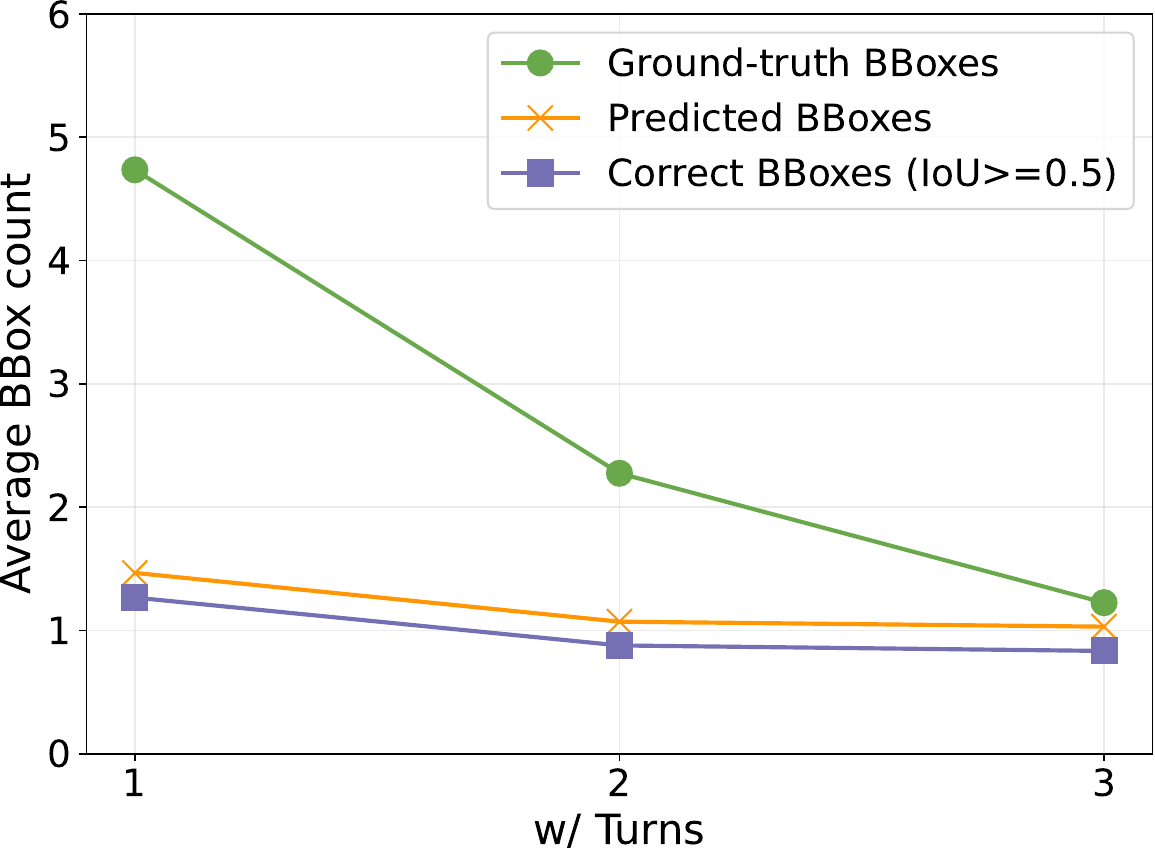}
        \subcaption{Qwen3.5 27B}
        \label{fig:ct_qwen35-27B}
    \end{subfigure}
    \begin{subfigure}[t]{0.32\linewidth}
        \centering
        \includegraphics[width=\linewidth]{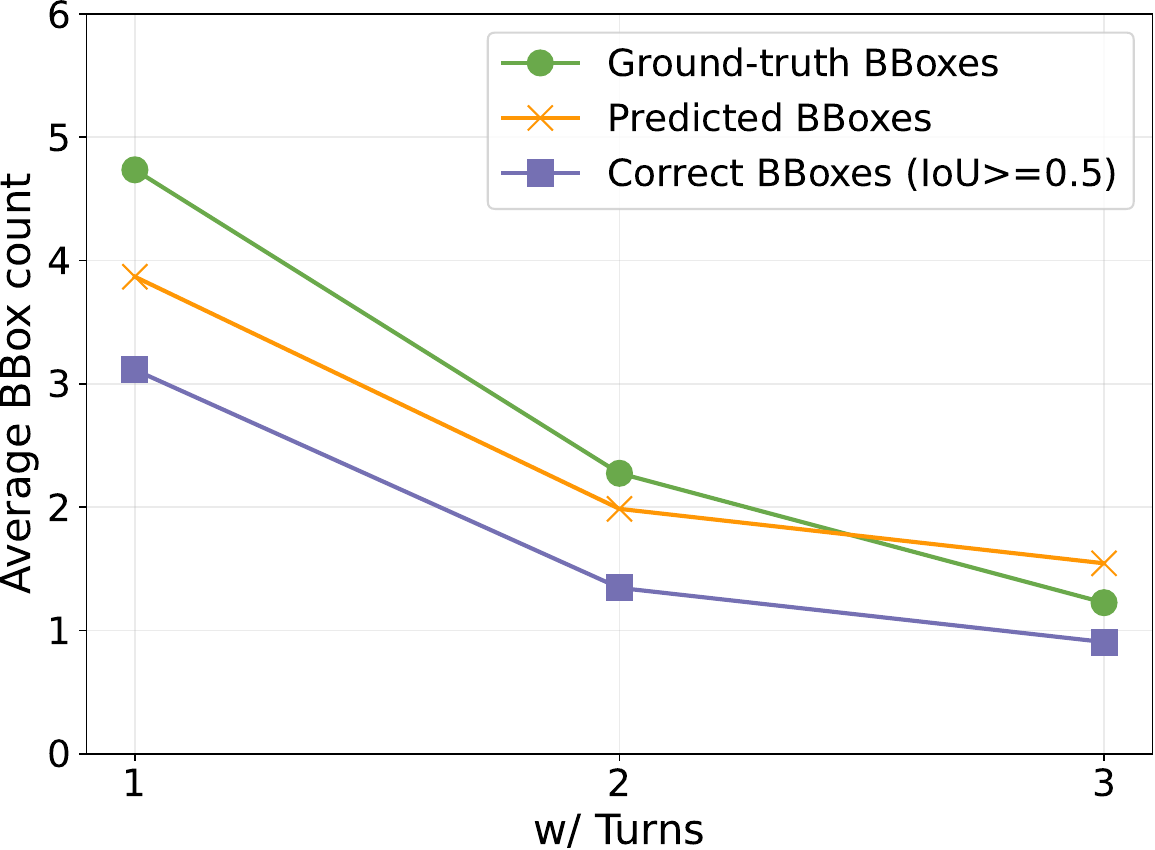}
        \subcaption{GPT-5.6 Luna}
        \label{fig:ct_gpt56-luna}
    \end{subfigure}
    \caption{Transitions of candidate sets across dialogue turns, showing the average numbers of ground-truth, predicted, and correctly predicted BBoxes. The analysis tracks the same 1,234 dialogues across the first three turns, covering both CSVG and CVG cases.}
    \label{fig:transition_cand-size}
\end{figure*}

\begin{figure*}[t]
    \centering
    \includegraphics[width=\linewidth]{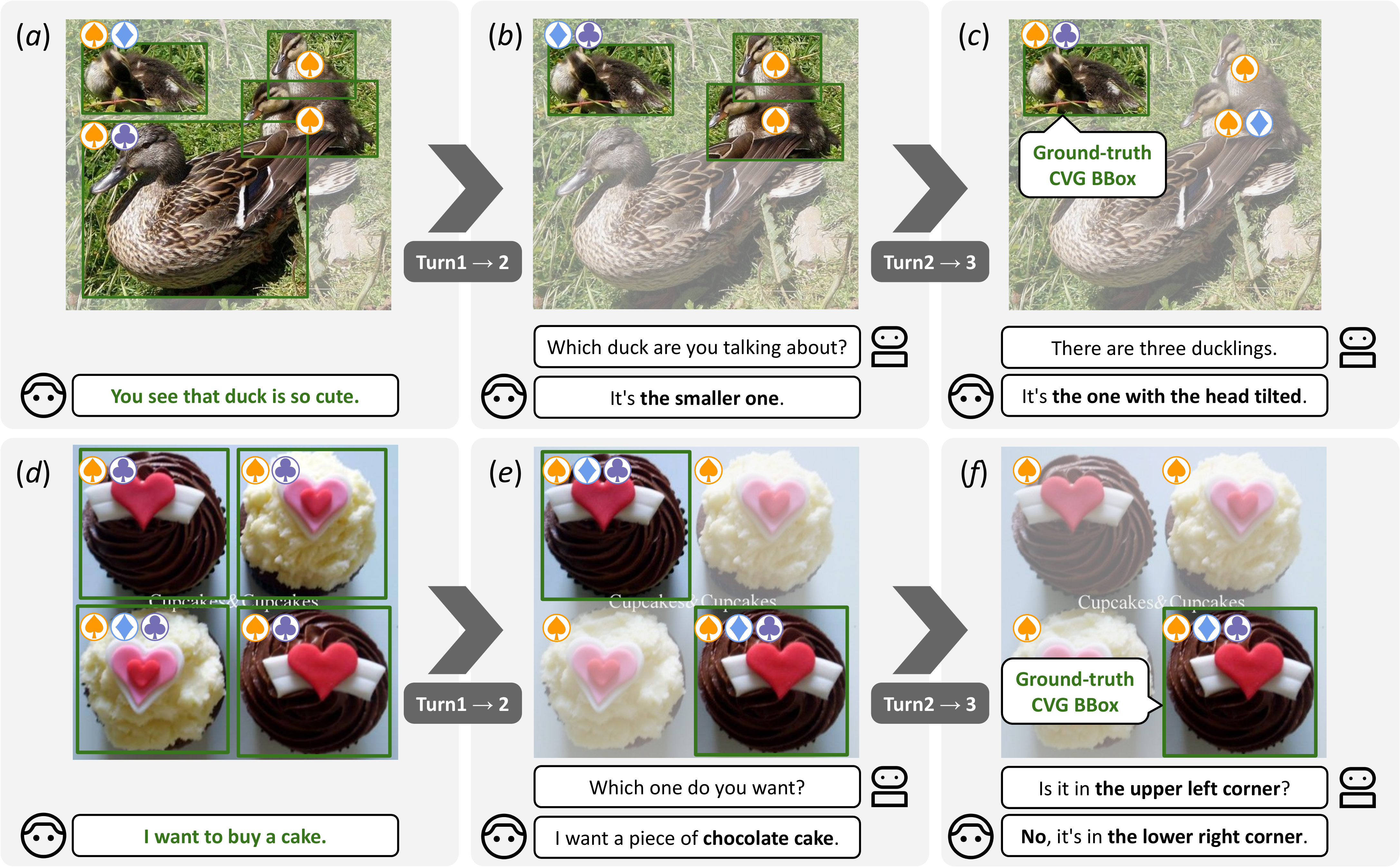}
    \caption{Examples of candidate set updates across turns: Green BBoxes indicate ground-truth candidates across both CSVG and CVG cases. The symbols {\color[HTML]{FF9D02}$\spadesuit$}, {\color[HTML]{6C9EED} $\blacklozenge$}, and {\color[HTML]{7671B5}$\clubsuit$} denote objects predicted by {\color[HTML]{FF9D02} \textbf{Qwen3 VL 4B}}, {\color[HTML]{6C9EED} \textbf{Qwen3.5 27B}}, and {\color[HTML]{7671B5} \textbf{GPT-5.6 Luna}}, respectively.}
    \label{fig:pred_examples}
\end{figure*}
\section{Analysis and Discussion}
The experimental results revealed that current VLMs have a large precision--recall gap on the CSVG task in CandVG, primarily due to low recall.
To understand the source of this gap, we analyze how candidate sets change as dialogue context accumulates and investigate whether VLMs can track these updates throughout a dialogue ({\S}~\ref{sec:6_tracking_candidate_set}).
Building on this analysis, we next examine whether these limitations can be mitigated through task-aware instruction, namely, additional instructions tailored to the CSVG task ({\S}~\ref{sec:6_task-aware-prompting}).

\subsection{Tracking and Updating Candidate Sets}
\label{sec:6_tracking_candidate_set}
\paragraph{Setup.}
We analyzed three models: Qwen3 VL 4B, which achieved high recall among the open-weight generalist VLMs; Qwen3.5 27B, which achieved the highest CVG accuracy; and GPT-5.6 Luna, which achieved the joint-highest CSVG F1 score.
To ensure a consistent comparison across turns, we track the same 1,234 dialogues across the first three turns, including both CVG and CSVG samples.
For each model, we report the average numbers of ground-truth, predicted, and correctly predicted BBoxes at each turn.

\paragraph*{Results.}
As shown in Figure~\ref{fig:transition_cand-size} (a) and (b), Qwen3.5 27B estimates fewer candidates than the ground-truth set from the first turn, contributing to its low recall.
Qwen3 VL 4B initially estimates a candidate set closer in size to the ground-truth set and achieves higher recall; however, its predictions do not shrink with the ground-truth set and instead increase from Turn 2 to Turn 3.
Taken together, \emph{Qwen3.5 27B tends to commit prematurely to a narrow candidate set, whereas Qwen3 VL 4B fails to narrow it sufficiently as context accumulates}.


Figure~\ref{fig:transition_cand-size} (c) shows that GPT-5.6 Luna tends to narrow the candidate set prematurely, while potentially retaining extra candidates as ambiguity is resolved.
In particular, it predicts fewer candidates than the ground-truth set in Turns 1 and 2, while its predictions exceed the ground-truth set at Turn 3.
These findings highlight failure modes that need to be addressed to improve candidate set estimation.\footnote{See Appendices~\ref{sec:appendix_additional_cvg} and~\ref{sec:appendix_transition_analysis} for additional analyses.}

\paragraph*{Qualitative Examples.}
Figure~\ref{fig:pred_examples} presents qualitative examples of candidate set updates across dialogue turns.
The examples include CSVG cases through Turn 2, where multiple objects remain plausible, and a CVG case at Turn 3, where the context identifies a single object.

As shown in Figure~\ref{fig:pred_examples} (a) and (d), Qwen3.5 27B  commits to a single object in the early turns, but reaches the correct CVG target once additional context becomes available in Figure~\ref{fig:pred_examples} (c) and (f).
In contrast, Qwen3 VL 4B initially retains multiple objects but fails to narrow the candidate set in the later turns.
GPT-5.6 Luna follows the ground-truth candidate set updates more closely overall, although it also commits prematurely in Figure~\ref{fig:pred_examples} (a).
Collectively, these examples reveal distinct model behaviors in recognizing and resolving ambiguity, demonstrating that \emph{CandVG enables both behaviors to be analyzed within the same dialogue}. 

\subsection{Effect of Task-Aware Instruction}
\label{sec:6_task-aware-prompting}
\paragraph{Setup.}
The base prompt was designed to provide consistent task instructions across benchmarks.
Here, we added CSVG-specific instructions intended to (i) provide a stepwise procedure~\cite{Wang-etal_2023_ACL} and (ii) discourage premature commitment.
We then evaluated their effects on CSVG precision and recall in 4B and 27B--38B open-weight generalist VLMs.
\begin{tcolorbox}[colframe=gray,fontupper=\scriptsize,title=\small (i) Task-Aware Instruction 1 (TAI-1)]
\small
1. Identify what kind of object is described by the text. 

2. Enumerate all plausible objects in the image. 

3. If the dialogue context following the text is provided, 
   remove objects that are contradicted by the context.
\end{tcolorbox}
\begin{tcolorbox}[colframe=gray,fontupper=\scriptsize,title=\small (ii) Task-Aware Instruction 2 (TAI-2)]
\small
When the reference is ambiguous, retain every visually plausible referent unless it is clearly ruled out by the marked text or the dialogue context (if provided).
\end{tcolorbox}

\paragraph{Results.}
Table~\ref{tab:task-aware-instruction} shows that both instructions improve recall for all 27B+ models, with TAI-1 yielding larger gains.
In contrast, the effects on 4B models are inconsistent, as recall improves only for Qwen3 VL under TAI-1.
These findings suggest that task-aware instructions, as a form of prompt engineering~\cite{Gu-etal_2023_vlm-prompting, Schulhoff-etal_2025_prompt-report}, \emph{improve CSVG recall in larger VLMs; however, their effects remain model-dependent and do not consistently extend to precision}.\footnote{See the Limitations section for further discussion.}

\begin{table}[t]
    \centering
    \begin{subtable}[t]{\linewidth}
        \centering
        \scalebox{0.86}{
        \begin{tabular}{l|rr|rr}
            \toprule
            \multirow{2}{*}{\textbf{Model}} & \multicolumn{2}{c|}{\textbf{4B}} & \multicolumn{2}{c}{\textbf{27B+}} \\
            & \textbf{TAI-1} & \textbf{TAI-2}
            & \textbf{TAI-1} & \textbf{TAI-2} \\
            \midrule
            Intern3.5 & $-0.058$ & $-0.021$ & \pldiff{$+0.041$} & $-0.022$ \\
            Qwen3 VL  & \pldiff{$+0.055$} & $+0.013$ & $+0.015$ & $+0.003$ \\
            Qwen3.5   & $-0.043$ & $-0.008$ & $+0.011$ & $-0.003$ \\
            \bottomrule
        \end{tabular}
        }
        \subcaption{Precision}
        \label{tab:task-aware-instruction-precision}
    \end{subtable}

    \vspace{1mm}

    \begin{subtable}[t]{\linewidth}
        \centering
        \scalebox{0.825}{
        \begin{tabular}{l|rr|rr}
            \toprule
            \multirow{2}{*}{\textbf{Model}} & \multicolumn{2}{c|}{\textbf{4B}} & \multicolumn{2}{c}{\textbf{27B+}} \\
            & \textbf{TAI-1} & \textbf{TAI-2}
            & \textbf{TAI-1} & \textbf{TAI-2} \\
            \midrule
            Intern3.5 & $-0.014$ & $-0.007$ & \pldiff{$+0.036$} & $+0.011$ \\
            Qwen3 VL  & \pldiff{$+0.043$} & $-0.006$ & \pldiff{$+0.097$} & \pldiff{$+0.042$} \\
            Qwen3.5   & $-0.040$ & $-0.046$ & \pldiff{$+0.087$} & \pldiff{$+0.057$} \\
            \bottomrule
        \end{tabular}
        }
        \subcaption{Recall}
        \label{tab:task-aware-instruction-recall}
    \end{subtable}
    \caption{Improvements and degradations in CSVG precision and recall from the base prompt after adding Task-Aware Instructions (TAI). \pldiff{Colored} values indicate improvements of at least 0.03 over the base prompt.}
    \label{tab:task-aware-instruction}
\end{table}

%% file: contents/7_conclusion.tex
\section{Conclusion}
This paper proposed Candidate Set Visual Grounding to evaluate whether models can track how the candidate set changes as dialogue context becomes available.
To support this evaluation, CandVG serves as a benchmark that pairs CSVG with CVG to capture candidate set updates across dialogue turns.
This design enables ambiguity recognition and resolution to be analyzed within the same dialogue.
Our evaluation of VLMs showed that model scale is associated with both stronger visual grounding performance and higher CVG accuracy, but not with consistently better CSVG F1 scores.
In CSVG, models often predicted only a subset of the candidates, resulting in a substantial precision--recall gap.
Our results showed that this gap is associated with difficulty in candidate generation and premature commitment, and that task-aware instructions improved recall for large-scale VLMs.
However, models with higher recall may still retain extra candidates even as ambiguity is resolved, suggesting that improving candidate set estimation requires addressing different failure modes.
Recognizing when an utterance remains ambiguous and which interpretations remain plausible is a prerequisite for systems that interact and collaborate naturally with users in the real world.
We believe that evaluating and improving this capability will advance the development of such systems.


%% file: contents/appendix.tex
\section{Data Collection Details}
\label{sec:appendix_data_collection_details}
CandVG is built upon InViG 21K~\cite{Zhang-etal_2024_CVPRW}, which contains natural human–human dialogues and provides BBoxes corresponding to the first utterance of each dialogue as well as BBoxes for objects in each image.
We use the former for preprocessing and data selection and as guidance during candidate set annotation, and the latter for mapping point annotations to BBoxes, as described below.

\paragraph{Preprocessing.}
We used all 2,998 dialogues from the validation and test splits of InViG 21K. 
Before annotation, we excluded dialogues where (i) fewer than two BBoxes corresponding to the first utterance were available, (ii) more than 15 such BBoxes were available, or (iii) the ratio between the largest and smallest BBox areas exceeded 20. 
We applied these criteria to ensure initial ambiguity, limit annotation cost, and reduce the influence of extremely large or small BBoxes in InViG, respectively, leaving 2,171 dialogues.

We then sampled 1,620 dialogues to balance the number of dialogues across bins defined by the number of unique BBoxes corresponding to the first utterance (2, 3, 4, 5, 6, and 7+).

\paragraph{Annotation Procedure.}
During the data selection stage described in {\S~\ref{sec:4_candvg_data_collection}}, we further excluded dialogues where (iv) the first utterance referred to objects absent from the image or did not refer to specific objects, (v) a single BBox covered a group of four or more objects, (vi) the BBoxes were invalid, or (vii) the referent of the first utterance was uniquely identifiable.

\begin{figure*}[t]
    \centering
    \includegraphics[keepaspectratio, width=0.95\textwidth]{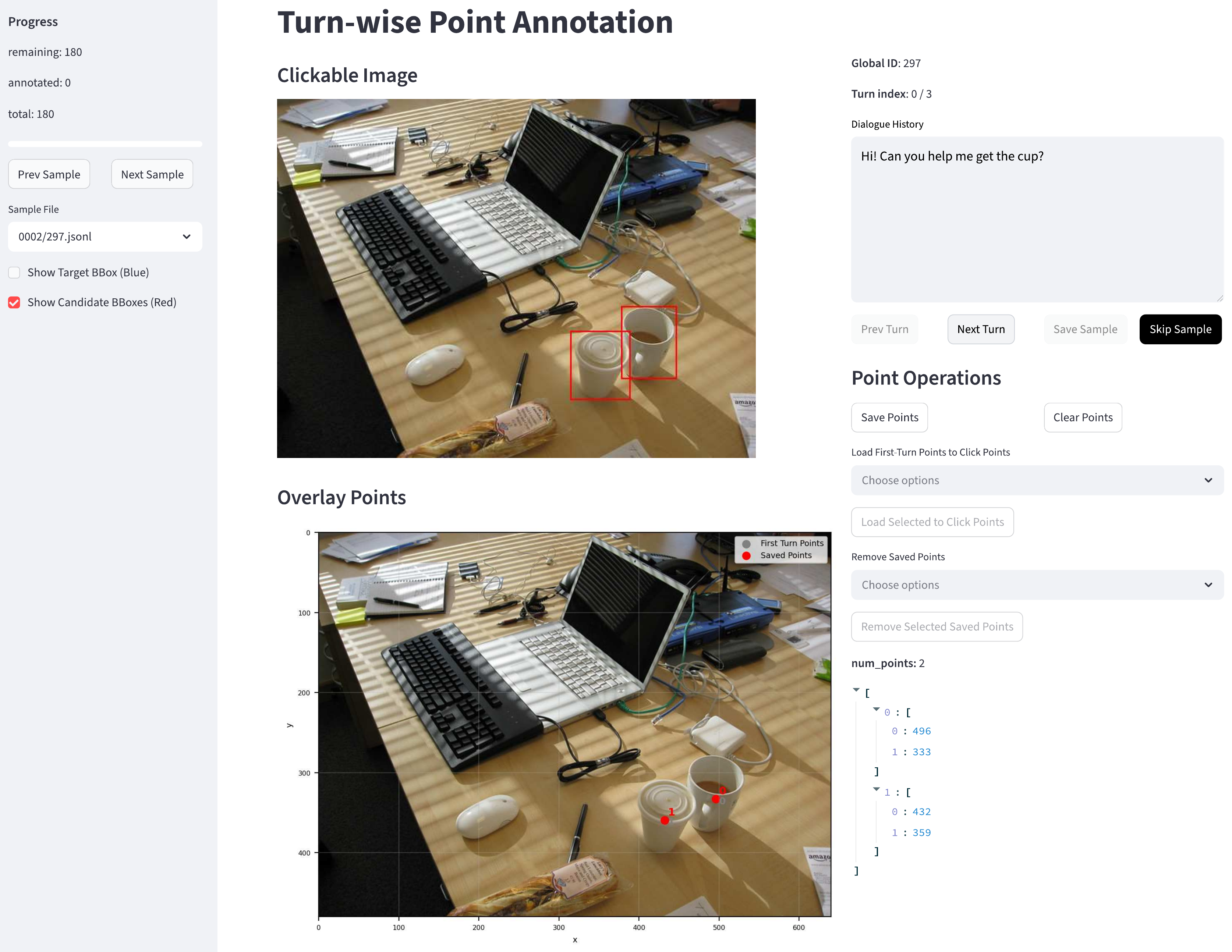}
    \caption{Annotation interface used for candidate set collection}
    \label{fig:annotation_tool}
\end{figure*}

Figure~\ref{fig:annotation_tool} shows the annotation interface used to collect candidate sets.
For annotation efficiency, annotators were shown the BBoxes corresponding to the first utterance provided by InViG, but their point annotations were not restricted to these BBoxes.
At each turn, annotators placed points on all objects they considered plausible given the image and the dialogue context, typically placing each point at the center of the corresponding BBox.\footnote{Point-based annotation has been used in prior work as an efficient annotation strategy for object localization~\cite{Papadopoulos-etal_2017_CVPR, Yu-etal_2022_CVPR}.}
When multiple BBoxes were assigned to the same object, they selected one of them and placed the point at its center.
When the BBox was difficult to identify visually, they instead placed the point at the center of the object itself.
The interface allowed annotators to add or remove points as the dialogue context was updated.

After annotation, we used the BBoxes provided by InViG for each image to map the annotated points.
Each point was mapped based on the center positions of these BBoxes.
We retained only dialogues for which a complete one-to-one correspondence between annotated points and unique BBoxes could be established at every turn.
This procedure resulted in the final 1,249 dialogues used in CandVG.

\begin{table}[t]
    \centering
    \scalebox{0.86}{
    \begin{tabular}{lr}
        \toprule
        Dialogues & 1,249 \\
        Samples & 4,833 \\
        Average  of Turns & 3.869 \\
        Average  of Objects & 2.355 \\
        Average of Unique Objects & 4.796 \\
        Total Words & 124,047 \\
        Unique Words & 2,058 \\
        \bottomrule
    \end{tabular}
    }
    \caption{Statistics on CandVG}
    \label{tab:basic_stats}
\end{table}
\section{Additional Statistics and Analysis}
\label{sec:appendix_additional_stats_analysis}
\begin{figure}[t]
    \centering
    \includegraphics[keepaspectratio, width=\linewidth]{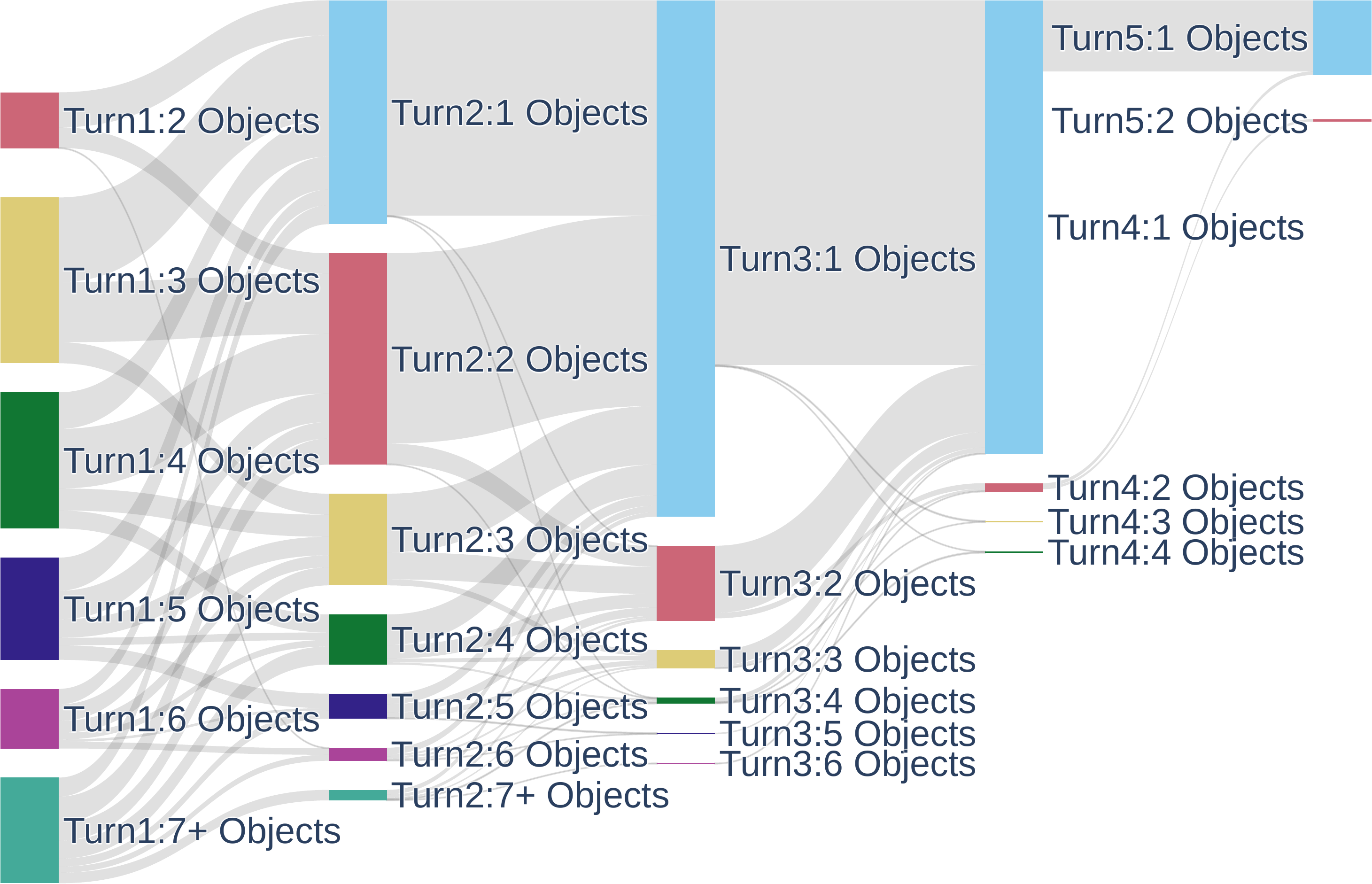}
    \caption{Transition of candidate set sizes across turns}
    \label{fig:transition_candidate}
\end{figure}

Table~\ref{tab:basic_stats} summarizes the basic statistics of CandVG. 
\paragraph*{Candidate Set Transition.}
Figure~\ref{fig:transition_candidate} illustrates transitions in candidate set size across dialogue turns.
Candidate sets become smaller as dialogue context accumulates, with many samples converging to a single object by the third turn.
This reflects the structure of the underlying InViG 21K, in which ambiguity in the first utterance is progressively resolved through subsequent dialogue context.\footnote{In 21 dialogues (1.68\%), multiple candidates remain plausible even at the final turn. We treat these final-turn samples as CSVG evaluation samples.}
After the third turn, subsequent utterances typically serve to confirm the identified object or provide non-referential acknowledgement.

\begin{figure*}[t]
    \centering
    \begin{subfigure}{0.32\linewidth}
        \centering
        \includegraphics[width=\linewidth]{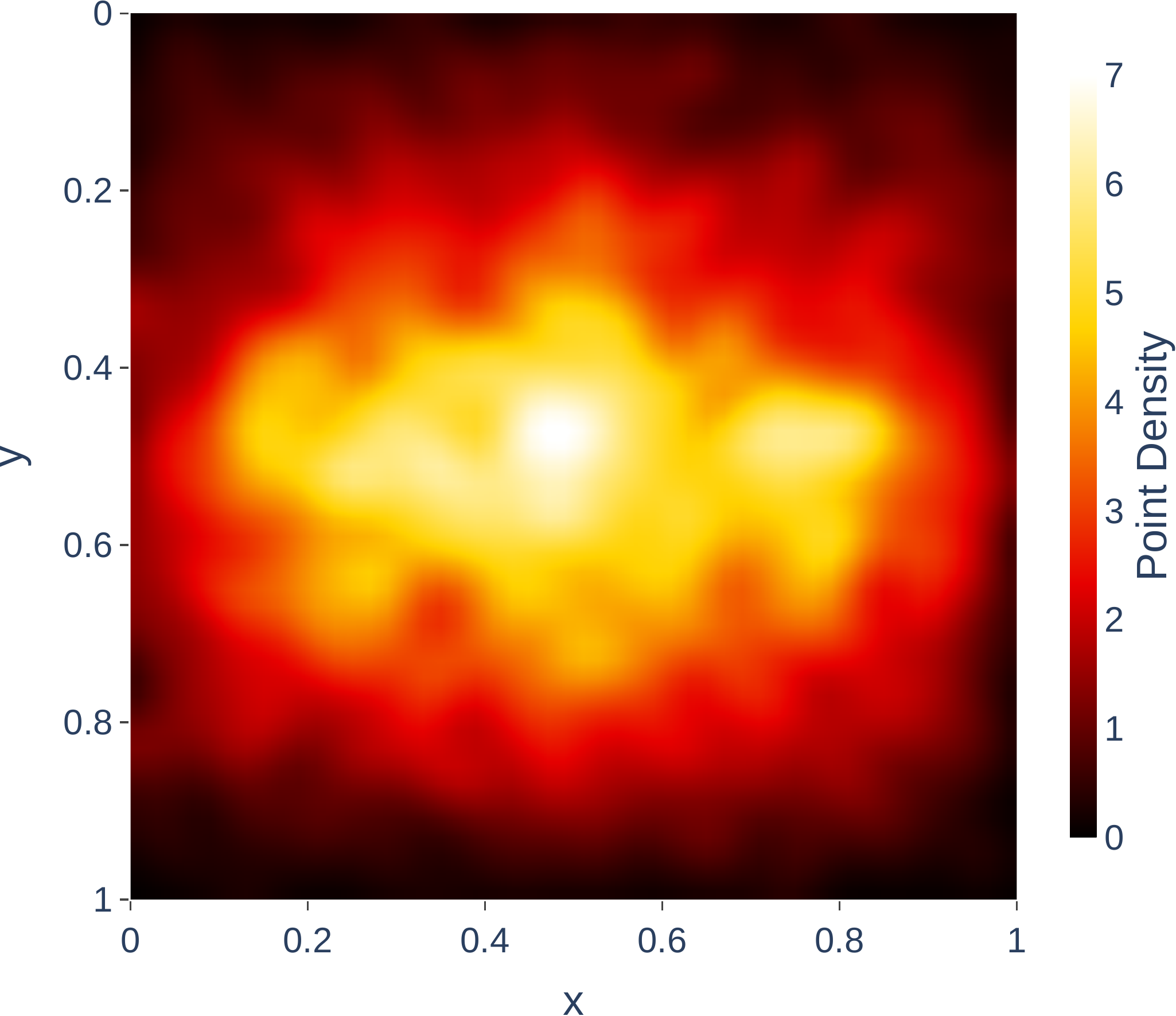}
        \caption{Turn 1}
        \label{fig:point_heatmap_turn0}
    \end{subfigure}
    \begin{subfigure}{0.32\linewidth}
        \centering
        \includegraphics[width=\linewidth]{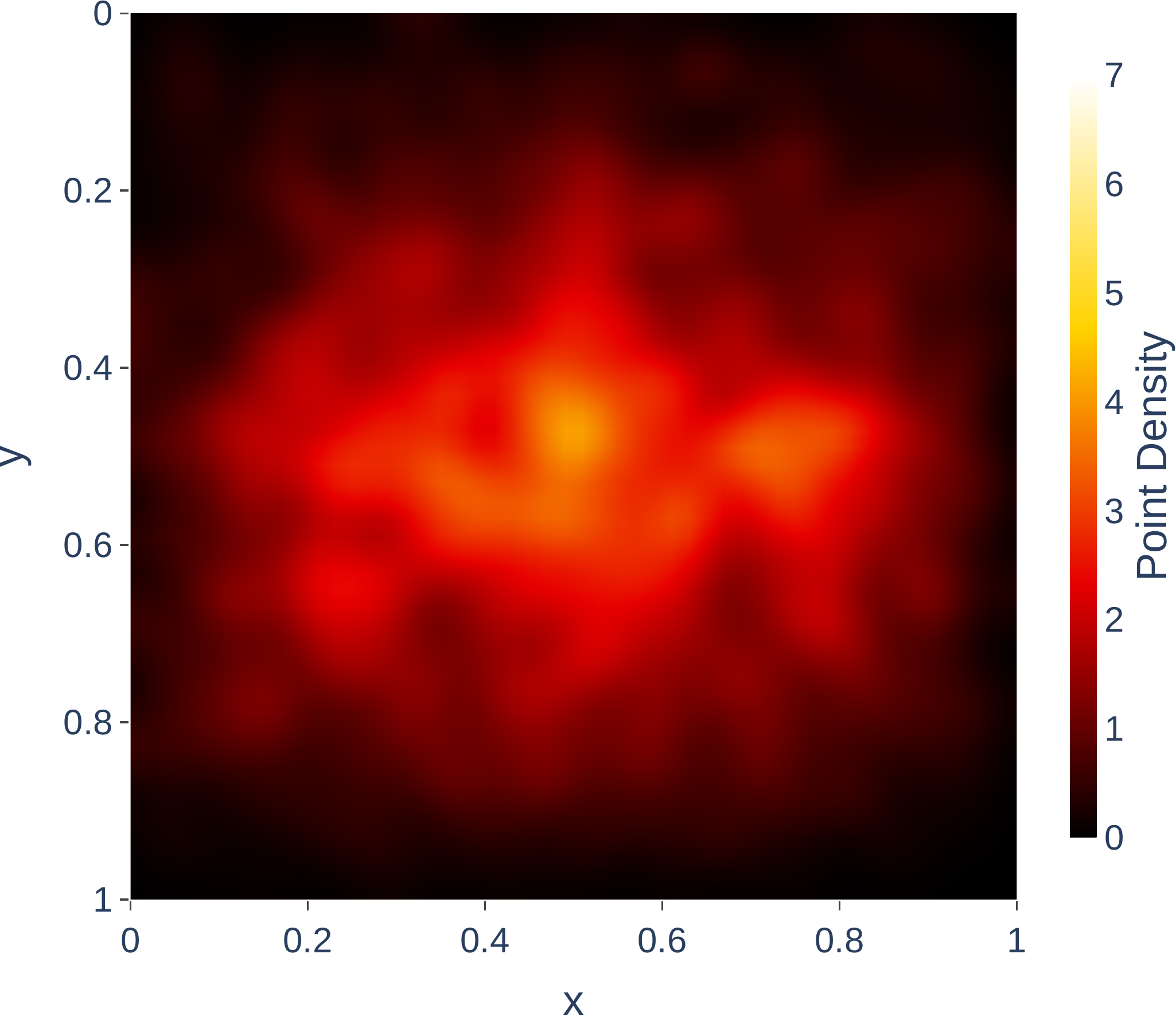}
        \caption{w/ Turn 2}
        \label{fig:point_heatmap_turn1}
    \end{subfigure}
    \begin{subfigure}{0.32\linewidth}
        \centering
        \includegraphics[width=\linewidth]{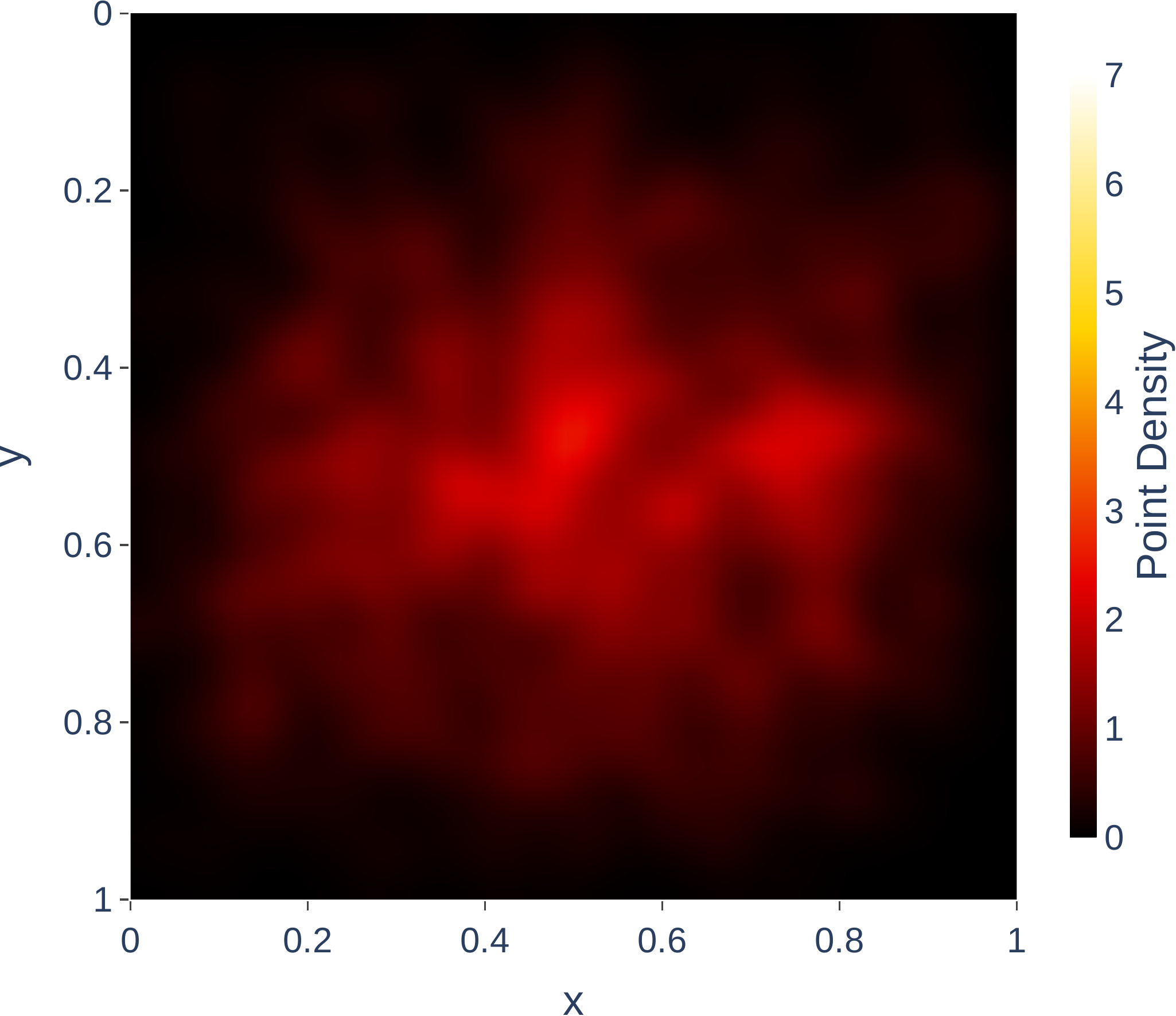}
        \caption{w/ Turn 3}
        \label{fig:point_heatmap_turn2}
    \end{subfigure}
    \caption{Heatmaps of point annotations across turns: Coordinates are normalized by image width and height.}
    \label{fig:point_heatmap}
\end{figure*}

\paragraph{Spatial Distribution of Point Annotations.}
Figure~\ref{fig:point_heatmap} shows heatmaps of point annotations across dialogue turns, illustrating the spatial distribution of annotated candidate objects.
To generate the heatmaps, we normalized point coordinates by the image width and height, aggregated them into two-dimensional histograms, and applied Gaussian smoothing.
The annotations are concentrated around the image center while still covering the entire image area.
The lower densities in later turns reflect the progressive reduction in candidate set size.

\section{Implementation Details}
\label{sec:appendix_impl_details}
Table~\ref{tab:model_cards} lists the HuggingFace model checkpoints for the open-weight models and the OpenAI API model cards for the proprietary models used in our experiments.
For the open-weight models, we used greedy decoding with a maximum generation length of 512 tokens.\footnote{Our preliminary experiments showed that greedy decoding performed better than beam search while producing nearly deterministic outputs.}
For GPT-5.6, we set the maximum generation length to 2,048 tokens and the reasoning effort to medium.
We ran all experiments with the open-weight models on four NVIDIA RTX 3090 GPUs, and the longest inference run for a single model took about six hours.

\begin{table}[t]
    \centering
    \scalebox{0.72}{
        \begin{tabular}{ll}
        \toprule
        \textbf{Models} & \textbf{HuggingFace ID or API Name} \\
        \midrule
        Intern3.5 VL 2B & \url{OpenGVLab/InternVL3_5-2B-HF} \\
        Intern3.5 VL 4B & \url{OpenGVLab/InternVL3_5-4B-HF} \\
        Intern3.5 VL 8B & \url{OpenGVLab/InternVL3_5-8B-HF} \\
        Intern3.5 VL 38B & \url{OpenGVLab/InternVL3_5-38B-HF} \\
        \addlinespace[4pt]
        Qwen3 VL 2B & \url{Qwen/Qwen3-VL-2B-Instruct} \\
        Qwen3 VL 4B & \url{Qwen/Qwen3-VL-4B-Instruct} \\
        Qwen3 VL 8B & \url{Qwen/Qwen3-VL-8B-Instruct} \\
        Qwen3 VL 32B & \url{Qwen/Qwen3-VL-32B-Instruct} \\
        \addlinespace[4pt]
        Qwen3.5 2B & \url{Qwen/Qwen3.5-2B} \\
        Qwen3.5 4B & \url{Qwen/Qwen3.5-4B} \\
        Qwen3.5 9B & \url{Qwen/Qwen3.5-9B} \\
        Qwen3.5 27B & \url{Qwen/Qwen3.5-27B} \\
        \addlinespace[4pt]
        GPT-5.6 Luna & \url{gpt-5.6-luna} \\
        GPT-5.6 Terra & \url{gpt-5.6-terra} \\
        \addlinespace[4pt]
        RynnBrain 2B & \url{Alibaba-DAMO-Academy/RynnBrain-2B} \\
        RynnBrain 8B & \url{Alibaba-DAMO-Academy/RynnBrain-8B} \\
        \addlinespace[4pt]
        CosmosReason2 2B  & \url{nvidia/Cosmos-Reason2-2B} \\
        CosmosReason2 8B  & \url{nvidia/Cosmos-Reason2-8B} \\
        \bottomrule
        \end{tabular}
    }
    \caption{Details of the models for the experiments}
    \label{tab:model_cards}
\end{table}

Table~\ref{tab:prompt_template} presents the base prompt and the model-specific BBox output formats used in the experiments.
For open-weight models, BBox coordinates are normalized to the range [0, 1000], whereas GPT-5.6 uses coordinates with respect to the original image width and height.

\begin{table*}[t]
\centering

\begin{tcolorbox}[colframe=gray,fontupper=\scriptsize,title=\small Base Prompt]
\begin{verbatim}
Provide the bounding box coordinates of the text described between <ref> and </ref>.
Return as box(es) as necessary, but include all plausible referents, up to a maximum of 15.

[Output Format]

<ref>Can you pass me the drink?</ref>
Which one is yours?
...

[Optional Candidate Selection Setting]
\end{verbatim}
\end{tcolorbox}

\begin{tcolorbox}[colframe=gray,fontupper=\scriptsize,title=\small Output Formats]
\begin{verbatim}
The placeholder [Output Format] is replaced by one of the following formats, with real 
coordinates filled in when a ground-truth candidate set is provided:

Qwen3 VL, Qwen3.5, CosmosReason2, and GPT-5.6:
[
{"bbox_2d": [x1, y1, x2, y2]},
...
]

Intern3.5 VL:
<box>[[x1, y1, x2, y2], ...]</box>

RynnBrain:
<object>(x1, y1), (x2, y2)</object> ...
\end{verbatim}
\end{tcolorbox}
\begin{tcolorbox}[colframe=gray,fontupper=\scriptsize,title=\small Candidate Selection Setting]
\begin{verbatim}
Candidate boxes (select from these boxes; coordinates are normalized to 0-1000):

[Output Format]
\end{verbatim}
\end{tcolorbox}

\caption{Prompts used in the experiments: All models share the same base prompt and differ only in output format. The selection setting optionally appends a ground-truth candidate set.}
\label{tab:prompt_template}
\end{table*}

\section{Additional Experiments}

\subsection{Results on CVG}
\label{sec:appendix_additional_cvg}

\paragraph{Failure to Narrow Candidate Sets.}
Acc@0.5, the primary metric for CVG, evaluates only the first predicted BBox and does not account for additional predictions ({\S}~\ref{sec:eval_metrics}).
To examine whether models retain extra candidates even when the object is uniquely identifiable, we additionally computed precision and recall using the same evaluation protocol as in the CSVG task.
In this setting, \emph{precision penalizes extra predictions}, while \emph{recall measures whether the ground-truth BBox is included among the predictions}.
We also computed the percentage of samples with two or more predicted BBoxes.\footnote{We do not report the percentage of samples with zero predicted BBoxes, since all models predicted at least one BBox for every CVG sample.}

As shown in Table~\ref{tab:cvg_set_metrics}, Qwen3 VL 4B predicts multiple BBoxes for 74.2\% of CVG samples, resulting in lower precision than recall.
This is consistent with its tendency to predict extra candidates as dialogue context accumulates, as shown in Figure~\ref{fig:transition_cand-size} (a).

Despite achieving comparable recall, GPT-5.6 models predict multiple BBoxes more frequently than Qwen3.5 27B, which may partly explain their lower CVG accuracy.
In particular, GPT-5.6 Luna predicts multiple BBoxes more frequently than Terra, consistent with its tendency to retain extra candidates as ambiguity is resolved, as shown in Figure~\ref{fig:transition_cand-size} (c).

\paragraph{Effect of Dialogue Context.}
We compared performance under three input conditions on the CVG task to verify that CandVG requires dialogue context beyond the first ambiguous utterance.
The conditions were the first utterance only (\emph{Turn 1}), the minimum context needed to uniquely identify the referred object (\emph{Necessary}), and the full dialogue context (\emph{Full}).
We evaluated the open-weight models using Acc@0.5 (Acc) as the CVG metric, computed at the dialogue level.

As shown in Table~\ref{tab:cvg_effect_context}, performance substantially improved from Turn 1 to Necessary across all models, confirming the importance of dialogue context for CVG in CandVG.
Performance under Necessary and Full was generally comparable, while Full achieved slightly higher accuracy for most models.
This suggests that later turns may provide additional descriptions that reinforce the referred object even after it becomes uniquely identifiable.

\begin{table}[p]
    \centering
    \scalebox{0.8}{
        \begin{tabular}{l|cccc}
        \toprule
        \textbf{Models}
        & \textbf{Acc ($\uparrow$)} 
        & \textbf{P ($\uparrow$)}
        & \textbf{R ($\uparrow$)} 
        & \textbf{$\geq$ 2 ($\downarrow$)} \\
        \midrule
        Intern3.5 VL 2B  & 0.521 & 0.519 & 0.533 & 2.9 \\
        Intern3.5 VL 4B  & 0.568 & 0.563 & 0.588 & 5.1 \\
        Intern3.5 VL 8B  & 0.580 & 0.575 & 0.616 & 11.0 \\
        Intern3.5 VL 38B & \pldiff{0.746} & \pldiff{0.744} & \pldiff{0.748} & \pldiff{0.7} \\
        \addlinespace[2pt]
        Qwen3 VL 2B  & 0.480 & 0.478 & 0.498 & \pldiff{3.2} \\
        Qwen3 VL 4B  & 0.622 & 0.328 & 0.763 & 74.2 \\
        Qwen3 VL 8B  & 0.655 & 0.543 & 0.753 & 39.3 \\
        Qwen3 VL 32B & \pldiff{0.762} & \pldiff{0.709} & \pldiff{0.809} & 18.2 \\
        \addlinespace[2pt]
        Qwen3.5 2B  & 0.606 & 0.599 & 0.626 & 5.1 \\
        Qwen3.5 4B  & 0.735 & 0.662 & 0.793 & 20.3 \\
        Qwen3.5 9B  & 0.790 & 0.765 & 0.797 & 5.0 \\
        Qwen3.5 27B & \pldiff{\textbf{0.819}} & \pldiff{\textbf{0.817}} & \pldiff{0.820} & \pldiff{0.6} \\
        \addlinespace[2pt]
        GPT-5.6 Luna & 0.743 & 0.737 & \textbf{0.826} & 15.3 \\
        GPT-5.6 Terra & \pldiff{\pldiff{0.780}} & \pldiff{0.774} & \pldiff{\textbf{0.824}} & \pldiff{8.3} \\
        \midrule
        \addlinespace[2pt]
        \multicolumn{5}{c}{\textbf{Embodied Reasoning VLMs} based on Qwen3 VL} \\
        \addlinespace[2pt]
        RynnBrain 2B & 0.599 & 0.599 & 0.599 & \pldiff{\textbf{0.0}} \\
        RynnBrain 8B & \pldiff{0.691} & \pldiff{0.691} & \pldiff{0.691} & \pldiff{\textbf{0.0}} \\
        \addlinespace[2pt]
        CosmosReason2 2B & \pldiff{0.537} & 0.530 & 0.586 & \pldiff{10.5} \\
        CosmosReason2 8B & 0.519 & \pldiff{0.402} & \pldiff{0.718} & 59.0 \\
        \bottomrule
        \end{tabular}
    }
    \captionof{table}{Results on the CVG task using Acc@0.5 (\textbf{Acc}), precision (\textbf{P}), recall (\textbf{R}), and the percentage of samples with two or more predicted BBoxes (\textbf{$\geq$ 2 [$\%$]})}
    \label{tab:cvg_set_metrics}
\end{table}

\begin{table}[p]
    \centering
    \scalebox{0.8}{
        \begin{tabular}{l|ccc}
        \toprule
        \textbf{Models} & \textbf{Turn1} & \textbf{Necessary} & \textbf{Full} \\
        \midrule
        Intern3.5 VL 2B & 0.265 & 0.519 & 0.510 \\
        Intern3.5 VL 4B & 0.308 & 0.560 & 0.560 \\
        Intern3.5 VL 8B & 0.277 & 0.581 & 0.561 \\
        Intern3.5 VL 38B & 0.297 & 0.722 & 0.755 \\
        \addlinespace[2pt]
        Qwen3 VL 2B & 0.172 & 0.467 & 0.469 \\
        Qwen3 VL 4B & 0.285 & 0.593 & 0.623 \\
        Qwen3 VL 8B & 0.273 & 0.630 & 0.653 \\
        Qwen3 VL 32B & 0.335 & 0.726 & 0.767 \\
        \addlinespace[2pt]
        Qwen3.5 2B & 0.299 & 0.584 & 0.606 \\
        Qwen3.5 4B & 0.326 & 0.700 & 0.742 \\
        Qwen3.5 9B & 0.323 & 0.762 & 0.803 \\
        Qwen3.5 27B & 0.340 & 0.791 & 0.835 \\
        \midrule
        \addlinespace[2pt]
        \multicolumn{4}{c}{\textbf{Embodied Reasoning VLMs} based on Qwen3 VL} \\
        \addlinespace[2pt]
        RynnBrain 2B & 0.282 & 0.577 & 0.598 \\
        RynnBrain 8B & 0.315 & 0.662 & 0.701 \\
        \addlinespace[2pt]
        CosmosReason2 2B  & 0.224 & 0.525 & 0.517 \\
        CosmosReason2 8B  & 0.251 & 0.518 & 0.509 \\
        \bottomrule
        \end{tabular}
    }
    \caption{Results on the CVG task under different dialogue context conditions}
    \label{tab:cvg_effect_context}
\end{table}

\subsection{Results on CSVG}

\paragraph{Detailed Results}
Table~\ref{tab:detailed_csvg_results} provides detailed results for Table~\ref{tab:4_overall_results}, Figure~\ref{fig:csvg_precision_recall}, and Figure~\ref{fig:candvg_gt}, including precision, recall, and F1 scores in the CSVG setting.
We do not report the w/ GT results for GPT-5.6 because its native BBox coordinate representation differs from that used by the open-weight models.

\paragraph{Results by Candidate Set Size.}
Table~\ref{tab:csvg_bins} reports F1 scores by candidate set size.
For many open-weight models, F1 scores tend to decrease as the candidate set size increases, likely because of lower recall.
In contrast, GPT-5.6 shows stable performance across candidate set sizes.

\begin{table*}[t]
    \centering
    \scalebox{0.86}{
        \begin{tabular}{l|ccc|ccc|ccc}
            \toprule
            \multirow{2}{*}{\textbf{Models}}
            & \multicolumn{3}{c|}{\textbf{CandVG (w/o GT)}}
            & \multicolumn{3}{c|}{\textbf{CandVG (w/ GT)}}
            & \multicolumn{3}{c}{\textbf{gRefCOCO}} \\
            & \makebox[1.1cm][c]{\textbf{P}} & \makebox[1.1cm][c]{\textbf{R}} & \makebox[1.1cm][c]{\textbf{F1}}
            & \makebox[1.1cm][c]{\textbf{P}} & \makebox[1.1cm][c]{\textbf{R}} & \makebox[1.1cm][c]{\textbf{F1}}
            & \makebox[1.1cm][c]{\textbf{P}} & \makebox[1.1cm][c]{\textbf{R}} & \makebox[1.1cm][c]{\textbf{F1}} \\
            \midrule

            \addlinespace[2pt]
            \multicolumn{10}{c}{\textbf{Generalist VLMs}} \\
            \addlinespace[2pt]

            Intern3.5 2B  & 0.683 & 0.230 & 0.329 & 0.855 & 0.319 & 0.431 & 0.704 & 0.324 & 0.419 \\
            Intern3.5 4B  & 0.715 & 0.258 & 0.357 & 0.895 & \pldiff{\textbf{0.787}} & \pldiff{\textbf{0.792}} & 0.798 & \pldiff{0.631} & \pldiff{0.674} \\
            Intern3.5 8B  & 0.689 & 0.230 & 0.331 & 0.863 & 0.654 & 0.661 & 0.763 & 0.407 & 0.496 \\
            Intern3.5 38B & \pldiff{0.776} & \pldiff{0.268} & \pldiff{0.382} & \pldiff{0.932} & 0.696 & 0.744 & \pldiff{\textbf{0.832}} & 0.578 & 0.646 \\

            \addlinespace[3pt]

            Qwen3 VL 2B  & 0.562 & 0.189 & 0.271 & 0.888 & 0.430 & 0.526 & 0.683 & 0.295 & 0.391 \\
            Qwen3 VL 4B  & 0.622 & \pldiff{0.538} & \pldiff{0.542} & 0.874 & 0.470 & 0.561 & 0.685 & 0.517 & 0.560 \\
            Qwen3 VL 8B  & 0.668 & 0.398 & 0.454 & \pldiff{0.907} & 0.484 & 0.581 & \pldiff{0.739} & 0.514 & 0.576 \\
            Qwen3 VL 32B & \pldiff{0.772} & 0.480 & \pldiff{0.543} & 0.874 & \pldiff{0.525} & \pldiff{0.614} & 0.716 & \pldiff{0.539} & \pldiff{0.589} \\

            \addlinespace[3pt]

            Qwen3.5 2B  & 0.728 & 0.325 & 0.410 & 0.905 & 0.320 & 0.445 & 0.773 & 0.422 & 0.507 \\
            Qwen3.5 4B  & 0.816 & \pldiff{0.413} & \pldiff{0.494} & 0.903 & \pldiff{0.596} & \pldiff{0.646} & \pldiff{0.819} & \pldiff{0.613} & \pldiff{0.672} \\
            Qwen3.5 9B  & 0.799 & 0.325 & 0.427 & 0.940 & 0.533 & 0.623 & \pldiff{0.818} & 0.409 & 0.510 \\
            Qwen3.5 27B & \pldiff{\textbf{0.838}} & 0.326 & 0.441 & \pldiff{\textbf{0.964}} & 0.491 & 0.601 & 0.745 & 0.409 & 0.497 \\

            \addlinespace[3pt]

            GPT-5.6 Luna  & 0.789 & \pldiff{\textbf{0.617}} & \pldiff{\textbf{0.649}} & --- & --- & --- & 0.745 & 0.717 & 0.721 \\
            GPT-5.6 Terra & \pldiff{0.819} & 0.599 & \pldiff{\textbf{0.649}} & --- & --- & --- & \pldiff{0.798} & \pldiff{\textbf{0.750}} & \pldiff{\textbf{0.763}} \\

            \midrule

            \addlinespace[2pt]
            \multicolumn{10}{c}{\textbf{Embodied Reasoning VLMs} based on Qwen3 VL} \\
            \addlinespace[2pt]

            RynnBrain 2B     & 0.727 & 0.233 & 0.342 & 0.786 & 0.238 & 0.354 & \pldiff{0.511} & \pldiff{0.196} & \pldiff{0.277} \\
            RynnBrain 8B     & \pldiff{0.753} & \pldiff{0.243} & \pldiff{0.356} & \pldiff{0.919} & \pldiff{0.298} & \pldiff{0.433} & 0.473 & 0.181 & 0.256 \\
            CosmosReason2 2B & \pldiff{0.653} & 0.235 & 0.326 & 0.875 & 0.387 & 0.482 & \pldiff{0.679} & 0.439 & 0.494 \\
            CosmosReason2 8B & 0.647 & \pldiff{0.455} & \pldiff{0.478} & \pldiff{0.900} & \pldiff{0.729} & \pldiff{0.742} & 0.642 & \pldiff{0.453} & \pldiff{0.497} \\

            \bottomrule
        \end{tabular}
    }
    \caption{Detailed precision (\textbf{P}), recall (\textbf{R}), and F1 score on CSVG: w/o GT and w/ GT denote the settings without and with ground-truth candidate sets, respectively.}
    \label{tab:detailed_csvg_results}
\end{table*}

\begin{table*}[t]
    \centering
    \scalebox{0.86}{
    \begin{tabular}{l|cccccc|c}
        \toprule
        \multirow{2}{*}{\textbf{Models}}
        & \multicolumn{7}{c}{\textbf{CandVG (w/o GT)}} \\
        & \makebox[1.1cm][c]{\textbf{2} \scriptsize(708)}
        & \makebox[1.1cm][c]{\textbf{3} \scriptsize(554)}
        & \makebox[1.1cm][c]{\textbf{4} \scriptsize(387)}
        & \makebox[1.1cm][c]{\textbf{5} \scriptsize(257)}
        & \makebox[1.1cm][c]{\textbf{6} \scriptsize(146)}
        & \makebox[1.1cm][c]{\textbf{7+} \scriptsize(232)}
        & \makebox[1.1cm][c]{\textbf{All} \scriptsize(2,284)} \\
        \midrule

        \addlinespace[2pt]
        \multicolumn{8}{c}{\textbf{Generalist VLMs}} \\
        \addlinespace[2pt]

        Intern3.5 2B  & 0.445 & 0.345 & 0.293 & 0.249 & 0.202 & 0.161 & 0.329 \\
        Intern3.5 4B  & 0.451 & 0.378 & \pldiff{0.330} & \pldiff{0.292} & \pldiff{0.250} & \pldiff{0.201} & 0.357 \\
        Intern3.5 8B  & 0.461 & 0.350 & 0.288 & 0.240 & 0.192 & 0.148 & 0.331 \\
        Intern3.5 38B & \pldiff{0.533} & \pldiff{0.405} & \pldiff{0.330} & 0.263 & 0.234 & 0.174 & \pldiff{0.382} \\

        \addlinespace[3pt]

        Qwen3 VL 2B  & 0.356 & 0.317 & 0.248 & 0.200 & 0.140 & 0.103 & 0.271 \\
        Qwen3 VL 4B  & 0.555 & \pldiff{0.595} & \pldiff{0.552} & \pldiff{0.522} & 0.412 & \pldiff{0.465} & \pldiff{0.542} \\
        Qwen3 VL 8B  & 0.546 & 0.487 & 0.432 & 0.376 & 0.326 & 0.299 & 0.454 \\
        Qwen3 VL 32B & \pldiff{0.599} & 0.568 & 0.533 & 0.508 & \pldiff{0.454} & 0.420 & \pldiff{0.543} \\

        \addlinespace[3pt]

        Qwen3.5 2B  & 0.465 & 0.434 & 0.406 & 0.359 & 0.357 & 0.278 & 0.410 \\
        Qwen3.5 4B  & 0.561 & \pldiff{0.510} & \pldiff{0.484} & \pldiff{0.453} & \pldiff{0.416} & \pldiff{0.362} & \pldiff{0.494} \\
        Qwen3.5 9B  & \pldiff{0.567} & 0.447 & 0.373 & 0.310 & 0.326 & 0.242 & 0.427 \\
        Qwen3.5 27B & \pldiff{0.565} & 0.471 & 0.398 & 0.354 & 0.305 & 0.249 & 0.441 \\

        \addlinespace[3pt]

        GPT-5.6 Luna  & 0.632 & \pldiff{\textbf{0.663}} & \pldiff{\textbf{0.656}} & \pldiff{\textbf{0.657}} & 0.644 & 0.648 & \pldiff{\textbf{0.649}} \\
        GPT-5.6 Terra & \pldiff{\textbf{0.646}} & 0.643 & \pldiff{\textbf{0.657}} & 0.644 & \pldiff{\textbf{0.667}} & \pldiff{\textbf{0.658}} & \pldiff{\textbf{0.649}} \\

        \midrule

        \addlinespace[2pt]
        \multicolumn{8}{c}{\textbf{Embodied Reasoning VLMs} based on Qwen3 VL} \\
        \addlinespace[2pt]

        RynnBrain 2B      & 0.464 & 0.381 & \pldiff{0.301} & 0.241 & 0.199 & 0.151 & 0.342 \\
        RynnBrain 8B      & \pldiff{0.498} & \pldiff{0.384} & \pldiff{0.299} & \pldiff{0.256} & \pldiff{0.205} & \pldiff{0.159} & \pldiff{0.356} \\
        CosmosReason2 2B  & 0.431 & 0.346 & 0.299 & 0.248 & 0.199 & 0.167 & 0.326 \\
        CosmosReason2 8B  & \pldiff{0.538} & \pldiff{0.501} & \pldiff{0.489} & \pldiff{0.411} & \pldiff{0.432} & \pldiff{0.328} & \pldiff{0.478} \\

        \bottomrule
    \end{tabular}
    }
    \caption{Detailed F1 scores for the CSVG task by candidate set size: Numbers in parentheses indicate the number of samples in each bin.}
    \label{tab:csvg_bins}
\end{table*}

\section{Additional Transition Analysis}
\label{sec:appendix_transition_analysis}
We quantitatively examine error patterns in tracking and updating candidate sets across dialogue turns.
This analysis considers both CVG and CSVG cases.
In particular, we focus on two types of candidate set transitions: \emph{retention}, where a BBox remains plausible across consecutive turns, and \emph{removal}, where a previously plausible BBox becomes implausible.
We exclude cases where new BBoxes are added to the candidate set in the subsequent turn, as they occur in only 1.03\% of consecutive turn pairs in CandVG.

\paragraph*{Setup.}
Let $\mathcal{B}^{GT}_i$ and $\mathcal{B}^{Pred}_i$ denote the ground-truth and predicted candidate sets at the $ i$-th turn within a dialogue, respectively.
We define the sets of retained ($\mathcal{R}_i$) and removed ($\mathcal{D}_i$) candidates between turns $i$ and $i+1$ as follows:
\begin{equation}
    \mathcal{R}_i = \mathcal{B}^{GT}_i \cap \mathcal{B}^{GT}_{i+1}, \quad
    \mathcal{D}_i = \mathcal{B}^{GT}_i \setminus \mathcal{B}^{GT}_{i+1}.
\end{equation}
As in the CSVG evaluation ({\S}~\ref{sec:eval_metrics}), we match predicted BBoxes to ground-truth BBoxes using one-to-one matching with an IoU threshold of 0.5.

We consider two success rates: retention success and removal success.
Throughout this analysis, $\sum_i$ denotes summation over all consecutive turn pairs across all dialogues.
The retention success rate measures the proportion of ground-truth retained candidates
correctly predicted at turns $i$ and $i+1$: 
\begin{align}
    \mathrm{Retention\,Success}
    &= \frac{
        \sum_i |\mathcal{R}_i \cap \mathcal{B}^{Pred}_i
        \cap \mathcal{B}^{Pred}_{i+1}|
    }{
        \sum_i |\mathcal{R}_i|
    }.
\end{align}
The removal success rate measures the proportion of ground-truth removed candidates
predicted at turn $i$ but not at turn $i+1$:
\begin{align}
    \mathrm{Removal\,Success}
    = \frac{
        \sum_i |(\mathcal{D}_i \cap \mathcal{B}^{Pred}_i)
        \setminus \mathcal{B}^{Pred}_{i+1}|
    }{
        \sum_i |\mathcal{D}_i|
    }.
\end{align}

A low removal success rate can result from either failing to remove a previously predicted candidate at turn $i+1$, or failing to predict it at turn $i$.
To distinguish these cases, we additionally consider two failure rates as follows:
\begin{align}
    \mathrm{Not\,Removed}
    &= \frac{
        \sum_i |\mathcal{D}_i \cap \mathcal{B}^{Pred}_i
        \cap \mathcal{B}^{Pred}_{i+1}|
    }{\sum_i |\mathcal{D}_i|}, \\
    \mathrm{Not\,Predicted}
    &= \frac{
        \sum_i |\mathcal{D}_i \setminus \mathcal{B}^{Pred}_i|
    }{\sum_i |\mathcal{D}_i|}.
\end{align}
Among ground-truth removed candidates, $\mathrm{Not\,Removed}$ measures the proportion incorrectly retained at turn $i+1$, whereas $\mathrm{Not\,Predicted}$ measures the proportion not predicted at turn $i$.
$\mathrm{Removal\,Success}$ and these two failure rates sum to 100\% when expressed as percentages.


\begin{table*}[t]
    \centering
    \scalebox{0.86}{
    \begin{tabular}{l|c|ccc}
        \toprule
        & \shortstack{\textbf{Retention [$\%$]} \\ \scriptsize(5,424)}
        & \multicolumn{3}{c}{
            \shortstack{\textbf{Removal [$\%$]} \\ \scriptsize(4,677)}
        
        } \\
        \cmidrule(lr){3-5}
        \textbf{Model}
        & \makebox[1.6cm][c]{\textbf{Success} ($\uparrow$)}
        & \makebox[1.6cm][c]{\textbf{Success}} ($\uparrow$)
        & \makebox[1.6cm][c]{\textbf{Not Rem.} ($\downarrow$)}
        & \makebox[1.6cm][c]{\textbf{Not Pred.}} ($\downarrow$) \\
        \midrule

        \addlinespace[2pt]
        \multicolumn{5}{c}{\textbf{Generalist VLMs}} \\
        \addlinespace[2pt]

        Intern3.5 2B  & 24.8 & 9.2 & 5.5 & 85.4 \\
        Intern3.5 4B  & 27.3 & 12.0 & 4.9 & \pldiff{83.1} \\
        Intern3.5 8B  & 27.2 & 9.6 & 4.4 & 86.1 \\
        Intern3.5 38B & \pldiff{33.8} & \pldiff{12.8} & \pldiff{3.4} & 83.8 \\

        \addlinespace[3pt]

        Qwen3 VL 2B  & 20.1 & 7.7 & \pldiff{5.5} & 86.9 \\
        Qwen3 VL 4B  & \pldiff{49.8} & 19.8 & 26.4 & \pldiff{53.9} \\
        Qwen3 VL 8B  & 41.7 & 18.2 & 11.4 & 70.3 \\
        Qwen3 VL 32B & 48.0 & \pldiff{31.1} & 9.6 & 59.2 \\

        \addlinespace[3pt]

        Qwen3.5 2B  & 33.4 & 17.7 & 7.0 & 75.3 \\
        Qwen3.5 4B  & \pldiff{43.6} & \pldiff{22.6} & 9.8 & \pldiff{67.7} \\
        Qwen3.5 9B  & 37.1 & 19.0 & 4.0 & 77.0 \\
        Qwen3.5 27B & 39.0 & 19.5 & \pldiff{\textbf{2.8}} & 77.7 \\

        \addlinespace[3pt]

        GPT-5.6 Luna  & \pldiff{\textbf{58.6}} & 44.8 & 11.8 & \pldiff{\textbf{43.4}} \\
        GPT-5.6 Terra & 55.9 & \pldiff{\textbf{46.6}} & \pldiff{8.7} & 44.6 \\

        \midrule

        \addlinespace[2pt]
        \multicolumn{5}{c}{
            \textbf{Embodied Reasoning VLMs} based on Qwen3 VL
        } \\
        \addlinespace[2pt]

        RynnBrain 2B     & 27.3 & 9.1 & 6.1 & \pldiff{84.8} \\
        RynnBrain 8B     & \pldiff{31.6} & \pldiff{9.9} & \pldiff{4.6} & 85.5 \\
        CosmosReason2 2B & 25.8 & 10.2 & \pldiff{6.0} & 83.8 \\
        CosmosReason2 8B & \pldiff{44.2} & \pldiff{15.7} & 20.6 & \pldiff{63.7} \\

        \bottomrule
    \end{tabular}
    }
    \caption{
        Candidate set retention and removal rates [$\%$] across consecutive dialogue turns: Removal rate outcomes are categorized as Success, Not Removed, and Not Predicted.
        Numbers in parentheses indicate the total counts of retained and removed candidates.
    }
    \label{tab:appendix_transition_analysis}
\end{table*}

\paragraph*{Results.}
Table~\ref{tab:appendix_transition_analysis} reports the candidate set retention and removal rates.
Across all models, the ``Not Predicted'' rate is higher than the ``Not Removed'' rate, indicating that many candidates removed from the ground-truth candidate set at turn $i+1$ were not predicted at turn $i$.
This finding is consistent with the difficulty of candidate
generation ({\S}~\ref{sec:5_effect-gt-cand}).

As shown in {\S}~\ref{sec:5_comparison-vg}, most generalist VLMs
achieve higher recall on gRefCOCO than on CandVG.
This suggests that the high ``Not Predicted'' rates are not only attributable to limitations in multi-object visual grounding.
These findings provide support for the premature commitment discussed in {\S}~\ref{sec:6_tracking_candidate_set}, where models predict narrow candidate sets while the utterance remains ambiguous.

Failure patterns vary across models.
For example, Qwen3 VL 4B achieves a high retention success rate among open-weight models but also has the highest ``Not Removed'' rate.
Similarly, GPT-5.6 Luna achieves the highest retention success rate, while its ``Not Removed'' rate of 11.8\% is consistent with its tendency to retain extra candidates as ambiguity is resolved ({\S}~\ref{sec:6_tracking_candidate_set} and {\S}~\ref{sec:appendix_additional_cvg}).